\RequirePackage{fix-cm}
\documentclass[smallcondensed,final]{svjour3}     % onecolumn (ditto)
\smartqed  % flush right qed marks, e.g. at end of proof
\usepackage{graphicx}
\usepackage{amsmath}
\usepackage{amssymb}
\usepackage[lined,boxed,commentsnumbered, ruled]{algorithm2e}
\usepackage{array}
\usepackage{url}
\usepackage{float}
\usepackage{color}
\usepackage{hyperref}
\usepackage{pdflscape}
\usepackage{subfigure}
\usepackage{booktabs}

\definecolor{citecol}{rgb}{0,0,0.5}
\hypersetup{urlcolor=citecol,linkcolor=black,citecolor=citecol,colorlinks=true}

\def\fpcomment#1{\textcolor{green}{FP: #1}}
\def\rev#1{\textcolor{black}{#1}}

\newcommand{\ie}{\textit{i.e.}, }

\newcommand{\NM}{r_{\mathit{prt}}}
\newcommand{\Mind}{r_{\mathit{ind}}}

\newcommand{\cover}{\mathit{cover}}
\newcommand{\supp}{\mathit{sup}}
\newcommand{\leverage}{\mathit{leverage}}
\newcommand{\esupp}{\mathit{ExpSupport}}
\newcommand{\comp}{\mathit{comp}}
\newcommand{\bsp}{\mathit{BSP}}
\newcommand{\fpr}[1]{\mathopen{}\left(#1\right)}
\newcommand{\brak}[1]{\left\langle#1\right\rangle}

\newcommand{\mset}[1]{\mathit{mset}(#1)}
\newcommand{\makeunique}[1]{\mathit{makeunique(#1)}}
\newcommand{\posset}[1]{\mathit{posset(#1)}}
\newcommand{\seq}[1]{\mathit{seq(#1)}}
\newcommand{\restore}[1]{\mathit{restore(#1)}}
\DeclareMathOperator{\mean}{mean}

\journalname{Data Mining and Knowledge Discovery}

\begin{document}
\sloppy

\title{SkOPUS}
\subtitle{Mining top-$k$ sequential patterns under leverage}


\author{Fran\c{c}ois~Petitjean \and Tao~Li \and Nikolaj~Tatti \and Geoffrey~I.~Webb}


\institute{
Francois Petitjean\at
              Faculty of Information Technology, Monash University\\
              \email{francois.petitjean@monash.edu}
\and
Tao Li \at
              School of Electronic and Information Engineering, Nanjing University of Information Science and Technology, 
              \email{litao@nuist.edu.cn} 
           \and
            Nikolaj Tatti\at
Department of Information and Computer Science,  Aalto University\\
              \email{nikolaj.tatti@aalto.fi}
              \and
              Geoffrey I. Webb\at
              Faculty of Information Technology, Monash University\\
              \email{geoff.webb@monash.edu}
\and
}

\date{Received: 14 December 2015 / Accepted: 23 May 2016 / Published online: 14 June 2016}

\maketitle

\begin{abstract}
\rev{This paper presents a framework for exact discovery of the top-$k$ 
sequential patterns under Leverage.} It combines (1) a novel definition of the expected
support for a sequential pattern ---~a concept on which most interestingness
measures directly rely~--- with (2) SkOPUS: a new branch-and-bound algorithm
for the exact discovery of top-$k$ sequential patterns under a given measure of
interest. 
%Our interestingness measure is based on comparing the pattern support with the average support of its sister patterns, obtained by permuting (to certain extent) the items of the pattern. 
\rev{Our interestingness measure employs the partition approach. A pattern is interesting to the extent that it is more frequent than can be explained by assuming independence between any of the pairs of patterns from which it can be composed.}
The larger the support compared to the expectation under independence, the more interesting is the pattern.
We build on these two elements to exactly extract the $k$ sequential patterns with highest leverage, consistent with our definition of expected support. 

%\noindent
We conduct experiments on both synthetic data with known patterns and real-world datasets; both experiments confirm the consistency and relevance of our approach with regard to the state of the art.

This article was published in Data Mining and Knowledge Discovery and is accessible at \url{http://dx.doi.org/10.1007/s10618-016-0467-9}.
\end{abstract}

\section{Introduction}
\label{sec:1}
%\emph{\underline{the introduction of this paper}}

Extracting interesting patterns from data is a core data mining task. 
This paper introduces a method to efficiently and exactly \rev{identify the top-$k$ patterns in} a \emph{sequential} database, using as the measure of interest {\em leverage\/} --- the difference between a pattern's observed and expected frequency. \rev{To define the expected frequency, we use a maximum-likelihood estimate under an assumption of independence between any pair of patters from which it can be composed. }

The notion of interestingness is at the core of this paper. In early work on pattern mining, patterns were considered \emph{interesting} if they appeared \emph{frequently} in data \cite{AgrawalEtAl:1993}. The underlying idea was that the fact that something happens often is useful information for the data practitioner. However, as a mature body of research has now shown \cite{RefSiebes2006,RefWebb2007,RefHan2007,RefWebb2008,RefBoley2009,RefHam2010,RefTatti2010,RefVreeken2011,RefMampaey2012,RefWebb2014,RefTew2014}, \emph{frequency} is often a poor proxy for \emph{interestingness}. 
One reason for this is that many patterns should be expected to be frequent in real data. 
For instance, in the traditional market basket use case, if 90\% of people buy apples and 90\% of people buy pears, then the pattern $\{apples,pears\}$ will be frequent even if the two events are completely independent. 
When handling large databases, this phenomenon creates a deluge of frequent but uninteresting patterns. This is especially problematic for real-world applications, because the most frequent patterns will often be well-known; it is less frequent interactions within the data that are most likely to provide novel insights.
%extracting the interesting patterns from the set of all the patterns with a frequency of more than, say, 1\% gets harder than finding a needle in a haystack. 

When tackling sequential databases, this issue becomes even more critical; %considering independence of the purchase behaviours between different visits to the market, 
patterns such as ``buying apples and subsequently buying pears'' will be extremely common. 

This is a significant problem in real-world data, even with relatively short sequences. As a simple demonstration of this, we report in Table~\ref{tab:intro} the five most frequent sequential patterns in the book ``The Adventures of Tom Sawyer'', with every sentence constituting a record in the database (the rest of the paper will make it clear how we have performed this extraction). %The resulting sequences are relatively short, with an average length of $10.7$. 
\begin{table}[H]
\centering
\caption{\label{tab:intro}The five most frequent sequential patterns in the \emph{Adventures of Tom Sawyer}.}
\begin{tabular}
{p{2.3cm}<{\centering}c}
\toprule 
pattern & $support$ \\
\midrule
$\langle and, and\rangle$ & 13\% \\[1pt]

$\langle and, to\rangle$ & 9.8\% \\[1pt]

$\langle to, and\rangle$& 9.1\% \\[1pt]

$\langle of, and\rangle$ & 8.6\% \\[1pt]

$\langle and, of\rangle$& 8.0\% \\[2pt] \bottomrule
\end{tabular}
\end{table}

We can observe that most frequent sequential patterns are not interesting, and simply correspond to the permutations of the most frequent word `and' and each of the two next most frequent words `to' and `of'. This has motivated several approaches to mining interesting sequential patterns, which we detail in Section~\ref{sec:2}. The above example illustrates two main points about the incompatibility of frequency as a proxy for interestingness in sequential pattern discovery: 
\begin{enumerate}
\item The main reason for the high frequency of all these patterns, is the frequency of the individual words. In English, it has been shown that one word in four comes from $\{the, be, to, of, and, a, in, that, have, I\}$ \cite{RefOECFactsEnglish}. It is then not surprising that a sentence would contain several occurrences of those. If the sequences are long enough, any sequence of independent events (which we believe will not often be of interest) will become frequent. In fact, for any pattern with probability $p>0$ of occurring at any time in the data, the probability of its repetition tends to 100\% as the length of the sequences in the database increases (this is even true for singletons).\footnote{If $p(\text{``buying apples''})>0$ and $p(\text{``buying pears''})>0$ and these events are independent, then the probability of observing their sequence in data follows: 
$\lim_{l\rightarrow \infty}p(\langle apples, pears\rangle) = 1$} This clearly demonstrates that frequency is not a good proxy for interestingness for sequential databases. 
\item Let us consider the second and third most frequent sequential patterns $\langle and, to\rangle$ and $\langle to, and\rangle$. These two patterns have similar frequency. This directly questions the relevance of frequency as an interestingness measure. For sequential patterns the order of items should be key to determining whether the pattern is interesting. If all the orderings of the terms in a sequential pattern are equally frequent then an unordered pattern, such as an itemset, captures all of the information about the potentially interesting regularity in the data.  We argue that the sequential pattern should be reserved for regularities that cannot be fully captured by unordered patterns.  In this case, we can see that with such simple patterns of length $2$, frequency ranks the two possible orderings of $\{and,to\}$ as close to equivalent. The question is then: "Should a pattern $\langle i_1, i_2\rangle$ be considered \emph{sequentially} interesting if it appears as often as  $\langle i_2, i_1\rangle$ in the database?" We argue below that this contradicts the natural definition of interestingness for sequential patterns.
\end{enumerate}

\noindent
In this paper, we make the two following contributions: 
\begin{enumerate}
\item Scoring: Expected support is the core element of standard pattern mining measures of interest, such as \emph{leverage} or \emph{lift} \cite{RefWebb2011}, because most measures of interest involve a comparison between the observed support of a pattern in the data and its expected support \cite{piatetsky:daapos}. We introduce a new definition for the expected support of a sequential pattern and present an algorithm for its computation. Following our motivation above, \rev{our approach tries to find a model that is local to the tested pattern by considering its re-orderings. }
\item Search: We introduce SkOPUS: a sequential extension of the branch-and-bound OPUS algorithm \cite{RefWebb2014}. SkOPUS can extract, exactly and efficiently, the $k$ sequential patterns with highest leverage, \textit{i.e.}, the $k$ patterns with the highest difference between their observed and expected support. Note that we will show that our algorithm is not limited to our definition of the expected support, and can be directly used to extract the top-$k$ patterns under any definition of the expected support. 
\end{enumerate}

Our paper is divided into five main sections. 
We present the related research in Section~2. 
In Section~3, we detail the proposed framework for the discovery of the top-$k$ patterns \rev{with highest leverage.}
In Section~4, we present the results of experiments conducted on both synthetic data for which we control what patterns are actually present, and on real-world datasets. 
We conclude the paper and present some future work in Section~5. 

%Before entering the details of this paper, we wish to make an important note: as the first paper specifically targeting exact discovery of the sequential patterns with highest leverage, we restrict our scope to datasets representing sequences of simple items (as opposed to sequences of itemsets). As we will discuss in the conclusion, the definition of interestingness measures for sequential patterns of itemsets raises further complex issues that fall beyond the scope of this paper. 

\section{Related work}
\label{sec:2}
We structure the related work around the two main elements that this paper addresses: discovery of \emph{sequential} and \emph{interesting} patterns. 

\subsection{Mining interesting non-sequential patterns}
There is a very mature body of research about non-sequential frequent pattern mining \cite{RefHan2007}. As raised in the introduction, it has now long been identified that the major issue is not whether we can derive the complete set of frequent patterns under certain constraints, but whether we can derive a compact but high-quality set of patterns that will be useful for most applications \cite{RefHan2007,RefBoley2009,RefTatti2010}. 

While the major focus has been on efficient discovery of frequent patterns, there is a growing body of research into identifying \emph{interesting} patterns. Several methods aim at finding the set of patterns that will best describe the dataset, using a variety of methods for scoring the set such as entropy \cite{RefMampaey2012} or information theoretic frameworks \cite{RefSiebes2006,RefVreeken2011}. Other approaches rather try to define measures of interest for patterns \cite{RefTew2014,Szymon2004}, and then perform the extraction of the most interesting patterns for different measures \cite{bayardo1999mining,webb2000efficient,geng2007choosing,RefHam2010,RefWebb2014}. \rev{We refer the reader to~\cite{aggarwal:14:fpm}, for a more complete survey on frequent pattern mining, and discovering interesting patterns.}

\subsection{Mining frequent sequential patterns}
Sequential pattern mining extends frequent pattern mining to sequential databases \cite{mabroukeh2010taxonomy,mooney2013sequential}. Real-world applications are numerous and include analysis of the purchase behaviour of customers over time, analysis of Web clickstreams, and study of biological sequences. Algorithms for mining frequent sequential patterns from sequential databases were first proposed in the 1990s \cite{RefAgrawal1995,RefMannila1995,RefMannila1997}. After these seminal papers, researchers quickly moved to the development of algorithms for the extraction of \emph{frequent} sequential patterns of higher complexity (sequences of itemsets) \cite{RefMasseglia1998,RefPei2001,RefHan2000,RefYan2006,RefRaissi2008,RefTzvetkov2003,RefFournier2013}.

Note that different researchers have used different definitions of support and that these directly influence the patterns that they extract. Some methods, like ours, consider the support as being the number of sequences that have the sequential pattern as a subsequence, while others consider the number of times that the pattern occurs in all the sequences of the dataset, multiple counting a sequence that embeds the subsequence multiple times. Some application domains and problems will benefit from one approach while others will from the other approach. Our techniques extend directly to the second definition, should such an approach be desirable, the primary change simply being in the counting of the support of a pattern and its sub-patterns.

\subsection{Mining interesting sequential patterns}
%Looking back at the chronologies of these last discoveries, it is not very surprising that the question of mining frequent sequences of itemsets took priority over finding interesting sequential patterns, because interestingness became the focus of non-sequential pattern mining techniques after those extensions. 
It is only in the last decade that extracting interesting sequential patterns has emerged as an important research topic. 

\noindent
We distinguish two main families of methods: 
\begin{enumerate}
  \item methods using information theoretic approaches, such as the Minimum Description Length, to score a set of patterns that best explain the dataset.
  \item methods defining an expected support for the patterns under a null model (or hypothesis), and using it to score and rank the patterns by comparison to the actual observed support of the pattern
\end{enumerate}

\paragraph{Minimum Description Length methods scoring a set of patterns}
In \cite{Tatti2012,Lam2014mining-compressing}, \rev{the authors propose a Minimum Description Length (MDL) approach for 
scoring a set of patterns} relative to a dataset as well as a heuristic
search method to construct the set. Here the idea is to score a pattern set
instead of a just individual patterns; these patterns should explain the data
well but at the same time be non-redundant. These goals are quantified, with
scores derived from MDL principles. 

\paragraph{Methods based on expected support}
Deriving an expected support under some null hypothesis is particularly
complex for sequential patterns. Several approaches have made independence
assumptions between the elements composing the sequential patterns. That is,
they use a null hypothesis that the data are generated by a 0-order Markov
model or stationary and independent stochastic process
\cite{Gwadera2005-independence,Tatti2009,RefGwadera2010,RefLowKam2013}.

Approaches to deriving expected support from Markov chains have been proposed
in \cite{Gwadera2005-markov,Castro2012significant}\rev{; in particular, the statistical significance of the extracted patterns is studied in \cite{Castro2012significant}. }More complex
Markov models have been studied in \cite{Achar2012}, with statistical
significance studied in \cite{Achar2015}. 

Tatti \cite{Tatti2014-compact-windows} takes a different approach to
interestingness and posits that for some applications, interesting patterns
will be the ones that occur in a short window. He then builds a model of the
expected length of a sequence for a pattern to occur, and compares it to the
actual one. 

Finally, in very recent work Tatti \cite{Tatti2015-partition} introduces an
approach that is the most related to ours. This aproach
takes inspiration from Webb's approach to finding interesting (non-sequential) itemsets \cite{Webb2010},
and builds the expected support by looking at different partitions of the
sequential pattern. It then uses the derived score to rank the strict episodes.
The key difference between our work in this paper and that in Tatti
\cite{Tatti2015-partition} is the information from which the expectation is
derived: Tatti \cite{Tatti2015-partition} uses item probabilities as well as the number of gaps in a subpattern
to derive the expectation whereas\rev{, in this work,} we compare the pattern against all alternative orderings of the same items. Among other technical
differences, Tatti \cite{Tatti2015-partition} defines its measure for strict episodes whereas we focus on
sequential patterns, that is, serial episodes. Finally, the way we measure the
difference between the observed support and expected support allows us to use a
monotonic bound, and essentially mine top-$k$ episodes without \rev{first generating} a candidate set,
whereas the previously discussed methods require \rev{that} a candidate set \rev{be first generated}, typically a
set of frequent patterns mined with a low threshold.

%\subsection{The gap we aim at filling}
%Focusing on the traditional sequential pattern case, we show how to define a principled model of the expected frequency that partitions on any subset of the tokens in a pattern, with respect to which we identify the highest leverage sequential patterns. 

\section{Extracting the Top-K interesting sequential patterns}
\label{sec:3}
We introduce our method in this section. 
We first introduce some notation. 
We then detail our model for the expected frequency as well as how to derive our interestingness measure from it --~leverage. 
Finally, we introduce our efficient search algorithm to extract the most interesting sequential patterns under these measures. 

Note that we have made all code freely available to allow independent confirmation and extensions of our work at \url{https://github.com/fpetitjean/Skopus}. 

\subsection{Definitions}
\label{sec:3.1}
%Given a data domain $\mathcal{A}$, let $\mathcal{I}=\{i_1,i_2,\ldots,i_{|A|}\}$ be a finite set of items.
%An itemset $X$ is a non-empty subset of $I$.
A \emph{sequence} $S$ over a finite set of items $\mathcal{I}$ is an ordered list $\langle s_1,\ldots,s_\ell\rangle$, with $s_i\in\mathcal{I}, 1\le{i}\le{\ell}$, where $\ell$ is the length of sequence $S$, denoted by $|S|=\ell$. A \emph{sequential database} $\mathcal{D}=\{S_1,S_2,\ldots,S_n\}$ of size $n$ is a multiset of $n$ sequences over $\mathcal{I}$, where each $S_i$ is a sequence. 

\begin{definition}[Sub-sequence]
A sequence $S'=\big<t_1,t_2,\ldots,t_{\ell}\big>$ is a \emph{sub-sequence} of the sequence $S=\big<s_1,s_2,\ldots,s_k\big>$, if there exists an index sequence
$1\le{r_1}<{r_2}<\cdots<{r_\ell}\le k$, such that $t_j=s_{r_j}$ for all $1\le{j}\le \ell$. In such a case, we write $s'\prec{s}$.
\end{definition}
%\gwcomment{We need to add composition and difference: $<a,b,c>\backslash<a,c>\, = \,<b>$.}
\begin{definition}[Head and Tail]
The \emph{head} of a sequence is its first element and the \emph{tail} is its
remaining elements. Given an item $a$ and a sequence $T$, we write $\brak{a \mid T}$
to mean a sequence starting with the head $a$ followed by the tail $T$.
For example, $\langle a\mid \langle b,c\rangle\rangle = \langle a,
b,c\rangle$ and  $\langle a\mid \langle \rangle\rangle = \langle a\rangle$.
\end{definition}

\begin{definition}[Cover]
The \emph{cover} of a sequence $S$ in a sequential database $\mathcal{D}$ is the set of records of which $S$ is a sub-sequence,
\[
\cover(S,\mathcal{D})=\{S_i: S_i\in \mathcal{D},S\prec{S_i}\}.
\]
Such a sequence is often called \emph{sequential pattern}. 
\end{definition}
\begin{definition}[Count]
The \emph{count} of a sequence $S$ in a sequential dataset $\mathcal{D}$ is number of records of which $S$ is a subsequence,
\[
\#(S,\mathcal{D})=|\cover(S,\mathcal{D})|.
\]
\end{definition}

\begin{definition}[Support]
The \emph{support} of a pattern $S$ is the proportion of the $n$ records in database $\mathcal{D}$ of which $S$ is a subsequence, 
\[
\supp(S, \mathcal{D})=|\cover(S,\mathcal{D})|/n.
\]
\end{definition}
For notational convenience, and when it is clear in the context, we will omit $\mathcal{D}$. 

\rev{Note that this paper focuses on sequential patterns and as such does not aim at extracting patterns of the type ``$A$, then either $B$ or $C$, and then $D$''. As mentioned in related work, such patterns refer to the more general class of episodes (see for example \cite{Tatti2015-partition}). Nor does it seek patterns of the form ``sequences that contain $A$ are more likely to contain $B$, irrespective of their order.'' Such patterns are already addressed by the substantial literature on itemset discovery.}

\subsection{When is a sequential pattern interesting? }
\label{sec:3.2}

We now introduce our model for the expected frequency of a given sequential pattern, as well as the interestingness measures that we derive from it. We start by providing an intuition for our framework and then introduce its formal definition. 

\subsubsection{From interestingness to expected support}
``When is a sequential pattern interesting?'' is the main question that we tackle in this paper. 
We have motivated in the introduction that, for example, it is unlikely that a pattern $\langle and,of\rangle$ would be interesting if the pattern $\langle of,and\rangle$ is also. 
\rev{It is unclear when both these patterns could be interesting except in a context where the non-sequential pattern $\{and,of\}$ holds, when the presence of $and$ increases the frequency of $of$ and vice versa, irrespective of order. Our philosophy is that such patterns are more succinctly captured as non-sequential patterns.}

The starting point of our approach to assessing interestingness is that a pattern should be  interesting if is not possible to explain its frequency by the frequency of its sub-patterns \cite{RefWebb2010}. $\langle a,b\rangle$ should only be interesting if its frequency is greater than can be explained just by the frequency of $\langle a\rangle$ and $\langle b\rangle$. $\langle a,b,c,d\rangle$ should not be interesting if its frequency can be explained by the frequency of any of its constituent sequential sub-patterns such as $\langle a,c\rangle$, $\langle b,d\rangle$, and $\langle a, c, d\rangle$. For example, if buying shoes is often followed by buying socks, and buying jeans is often followed by buying a belt, then the pattern $\langle \text{shoes}, \text{jeans}, \text{socks}, \text{belt} \rangle$ should only be interesting if it is more frequent than should be expected given the frequency of $\langle \text{shoes}, \text{socks} \rangle$ and $\langle \text{jeans}, \text{belt} \rangle$ (as well as any other sub-sequences of this 4-element sequence).

Following the standard approach for defining interestingness measures, we assess interestingness in terms of deviation between observed support and expected support.  

\rev{An intuitive first approach to deriving the expected support might be to simply multiply the supports of the two subsequences. If $\langle a\rangle$ and $\langle b\rangle$ both occur in 50\% of sequences then perhaps $\langle a,b\rangle$ should be expected in 25\%. However, this ignores that they may also occur in the order $\langle b,a\rangle$, suggesting that there is a need to adjust for the number of permutations that could occur. But there is a further complication --- $\langle a\rangle$ and $\langle b\rangle$ can both occur in sequences of length 1 and $\langle a,b\rangle$ cannot. The length of the sequences in which the pattern appears must be taken into account, but exactly how this will affect the expectation is greatly dependent upon the type of distribution from which the data are drawn. It is far from clear what simple models might take account of all these factors without making very strong assumptions about the form of the distribution. We seek to develop a simple null hypothesis which can be tested without making any such assumptions.}

\subsubsection{Our proposed definition of expected support for sequential patterns}
Before developing the general formulas for patterns of any length, it is instructive to consider patterns of limited length to gain intuition about our general definitions. 

\paragraph{Pattern of length 2}
\iffalse
For a sequential pattern $\langle a,b\rangle$, we consider modifying the order of either sub-sequence $\langle a\rangle$ or $\langle b\rangle$. Because moving either $\langle a\rangle$ or $\langle b\rangle$ results in pattern $\langle b,a\rangle$, we define the expected support as 
\[
\esupp\fpr{\langle a,b\rangle} = \frac{\supp\langle a,b\rangle + \supp\langle b,a\rangle}{2}.
\]
This will directly lead to considering a pattern $\langle and,of\rangle$ interesting if and only if its support is different to the average support over all the possible orderings of the elements composing the pattern, that is, if and only if the order in which the events occur in the pattern matters in explaining its frequency (which is exactly our aim). 
\fi
\rev{We seek sequential patterns, that is, patterns in which the order of the items is significant.  That is, we want a pattern $\langle a,b\rangle$ to indicate that the presence of an $a$ in a sequence increases the probability of a $b$ appearing subsequently. We want to distinguish such a pattern from a co-occurrence pattern, $\{a,b\}$ which indicates that a sequence that contains an `$a$' is more likely to contain a `$b$', irrespective of the order of their appearance.  One simply hypothesis to test, which makes no assumptions about the form of the distribution from which the data are drawn, is that of all sequences in which $a$ occurs, there are more in which $b$ occurs after $a$ than there are sequences in which $b$ occurs before $a$. One way of stating this hypothesis is that $\supp\langle a,b\rangle>\supp\langle b,a\rangle$. If a statistical hypothesis test were developed, the null hypothesis would be that $\supp\langle a,b\rangle\leq\supp\langle b,a\rangle$.}

\paragraph{Patterns of length 3 or more}
\iffalse
For a sequential pattern $\langle a,b,c\rangle$,
we consider all the partitions created by a subsequence of the pattern and its
remainder (\textit{e.g.} $ (\langle a \rangle,\langle b,c\rangle)$),
and then
consider all other patterns that could be formed from these pairs while
respecting the order in both sequences:
\[
	\comp(\brak{x}, \brak{y, z}) = \{\brak{x, y, z}, \brak{y, x, z}, \brak{y, z, x}\},
\]
Note that $\comp$ should be red 'compositions' and will be formally defined in Definition~\ref{def:comp}. 
We we use this to define the expected support as:
\[
\begin{split}
\esupp\fpr{\brak{a,b,c}} = \max(&\mean\{\supp(S) \mid S \in \comp(\brak{a}, \brak{b, c})\} \\
&\mean\{\supp(S) \mid S \in \comp(\brak{b}, \brak{a, c})\} \\
&\mean\{\supp(S) \mid S \in \comp(\brak{c}, \brak{a, b})\}\,), 
\end{split}
\]
where $\mean$ is the arithmetic mean. 

A pattern of length 3 will thus be considered interesting if it appears significantly more frequently in the database than recompositions of this pattern that consider changing the position of $a$, $b$ and $c$, in turn. 
Taking the maximum over the three pairs of subsequences allows us to consider a pattern interesting if and only if its support is different to \emph{any} re-ordering with regard to a partition into subsequences. 
Taking the mean over the compositions\footnote{This term will be formally defined below. } constructs an expected support where each composition of the two subpatterns is considered equiprobable. 
\fi
\rev{The considerations become more complicated when we consider patterns of length three. One possibility would be that we want $\supp\langle a,b,c\rangle$ to be greater than the support of every other permutation of $a$, $b$ and $c$. However, that would disallow the possibility of both $\langle a,b,c\rangle$ and $\langle a,c,b\rangle$ being interesting patterns.  While we want to disallow all permutations of a set of items being interesting sequential patterns, there is no reason to mandate that only one permutation should be interesting.}

\rev{It is well understood in the context of non-sequential patterns, that unless specific steps are taken to prevent patterns from being accepted that are the composition of multiple interesting sub-patterns, or the inclusion of frequent singletons into an interesting pattern, the results of a pattern discovery system will often be swamped by myriads of spurious byproducts of the core patterns \cite{RefWebb2010}. One effective approach to prevent this is to test a pattern $X$ against every partition $Y,Z$ such that $Y\subsetneq X$ and $Z=X\backslash Y$. The counterpart of such a test for sequential patterns would test against every pair of subsequences from which it can be composed. Thus, we want to test $\langle a,b,c\rangle$ against $\langle a,b\rangle, \langle c\rangle$; $\langle a,c\rangle, \langle b\rangle$; and $\langle b,c\rangle, \langle a\rangle$. For the `partition' $\langle a,b\rangle, \langle c\rangle$ we can compose three sequential patterns by interweaving the two patterns, $\langle c,a,b\rangle$; $\langle a,c,b\rangle$; and $\langle a,b,c\rangle$. There are two obvious criteria that might be imposed to determine whether $\sup\langle a,b,c\rangle$ can be explained by $\sup\langle a,b\rangle$ and $\sup\langle c\rangle$. One is that $\sup\langle a,b,c\rangle$ must be greater than at least one of the supports of the other compositions of the generator.  The other is that $\sup\langle a,b,c\rangle$ must be greater than the mean of $\sup\langle c,a,b\rangle$; $\sup\langle a,c,b\rangle$; and $\sup\langle a,b,c\rangle$.  Both approaches are credible. The latter is a stronger constraint than the former, but still allows two of the three patterns that can be composed from a two-element and one-element sequential pattern to be found to be `interesting'.  In the current work we choose this stronger approach, but our software supports both. We thus assess $\langle a,b,c\rangle$ to be `interesting' if its support is greater than mean of the supports of the recompositions of every one of its sequential partitions.}

We now turn to the general formula for the expected support of a sequential pattern, for which we first introduce the notions of \emph{binary sequential partition} and of \emph{composition of two sequences}. 

%\iffalse
\begin{definition}[Sequential compositions]\label{def:comp}
\rev{The \emph{sequential compositions} of sequences $S$ and $T$, denoted $\comp(S, T)$,} is the set of
all sequences that contain all and only the elements of $S$ and $T$, respecting
their order. More formally, we define the composition recursively
\[
\begin{split}
\comp(S,\langle\rangle) &= \{S\}\\
\comp(\langle\rangle, T) &= \{T\}\\
\comp(\langle s_1 \mid S^*\rangle, \langle t_1 \mid T^*\rangle)& =
\{\langle s_1\mid U\rangle \mid U\in \comp(S^*, \langle t_1\mid T^*\rangle)\} \nonumber\\
&\qquad\cup \{\langle t_1\mid V\rangle \mid V\in \comp(\langle s_1\mid S^*\rangle, T^*)\}.
\end{split}
\]
% For example, 
% \[
% \begin{split}
% &\comp(\langle a, b\rangle,\langle c, d\rangle)\\
% &\qquad=\left\{\langle a, b, c, d\rangle,\langle a, c, b, d\rangle,\langle a, c, d, b\rangle,\langle c, a, b, d\rangle,\langle c, a, d, b\rangle,\langle c, d, a, b\rangle\right\}
% \end{split}
% \]
% but $\langle d,c,b,a\rangle\notin \comp(\langle a, b\rangle,\langle c, d\rangle)$.
\end{definition}

\begin{definition}[Binary sequential partition]
Let $S$, $S_1$ and $S_2$ be three sequences. We call $\{S_1,S_2\}$ a binary sequential partition of $S$, if $|S_1|>0$, $|S_2|>0$ and $S\in comp(S_1, S_2)$.
We denote as $\bsp(S)$ the set of all binary sequential partitions of $S$,
\[
\bsp(S)=\left\{\{S_1,S_2\} \mid S\in \comp(S_1, S_2)\right\}.
\]
For example, \rev{the binary sequential partitions of }
%\rev{\begin{multline*}
%\comp(\langle a,b\rangle, \langle c,d\rangle)=\\
%\left\{\langle a,b,c,d\rangle,\langle a,c,b,d\rangle, \langle c,a,b,d\rangle,\langle a,c,d,b\rangle,\langle c,a,d,b\rangle, \langle c,d,a,b\rangle\right\},
%\end{multline*}}
\begin{multline*}
\bsp(\langle a,b,c,d\rangle)=\left\{\{\langle a\rangle,\langle b,c,d\rangle\},\ \{\langle a,b\rangle,\langle c,d\rangle\},\ \{\langle a,b,c\rangle,
\langle d\rangle\},\right.\\
\left.\{\langle a,b,d\rangle,
\langle c\rangle\},\{\langle a,c\rangle,\langle b,d\rangle\},\ \{\langle a,c,d\rangle,\langle b\rangle\},\ \{\langle a,d\rangle,\langle b,c\rangle\}\right\}.
\end{multline*}

\end{definition}

We can now introduce our general definition for the expected support of a sequential pattern. 
\begin{definition}[Expected support for a sequential pattern]
Let $S$ be a sequential pattern, we define the expected support of $S$ as 
\[
\esupp(S)=\mathop{\max}_{(S_1,S_2)\in \bsp(S)}\lbrace\mathop{\mean}_{S'\in \comp(S_1,S_2)}\lbrace \supp(S')\rbrace\rbrace.
\]
\end{definition}
\rev{Let us give an intuition about this definition. We consider all the possible pairs of subsequences that respect the order in the target sequence (all the binary sequential partitions). Each of these partitions corresponds to a potential generator of the targeted sequence. For example, to determine the interestingness of the sequence $S=\langle a, b, c, d\rangle$, we look at how much more frequent it is than the maximum likelihood expectation if the data was generated from, say, $U=\langle a, d\rangle$ and $V=\langle b, c\rangle$ assuming that all possible ways to interweave $U$ and $V$ are equiprobable. We then take the maximum expected support over all potential binary sequential partitions (i.e. over all possible generators). This is because finding one pair $(U,V)$ that `explains' the support $S$ should be enough to establish that pattern $S$ is not interesting. 
It is important to note that we do not need to consider the expected support of compositions of more than two subsequences, as the relevant pairwise partitions will subsume compositions of more than two subsequences.
The aim of this paper is to extract the top-$k$ sequential patterns under \emph{leverage}, i.e.\ that will have maximum difference between their observed and expected support:
\[
	\leverage = \supp(S)-\esupp(S).
\]
A pattern will be considered interesting if its support cannot be explained by any of its sub-patterns.
%It is interesting to note that using leverage with our definition of expected support resembles the definition of \emph{improvement} for association rules \cite{Bayardo2000}.
}

\rev{It is finally important to note that if a pattern is considered interesting under our framework, then some of its sub-patterns might also be considered interesting. This will for example be the case with singular patterns that have low expected support. An example is the pattern $\langle if,you,can\rangle$ present in the poem ``If---'' by R. Kipling: it is present in more than 1/3 of the verses while none of its reorderings are, and also none of the re-orderings of its sub-patterns. For example the pattern $\langle can,you\rangle$, which is a reordering of $\langle you, can\rangle\prec\langle if,you,can\rangle$, never occurs either, thus making the pattern $\langle you, can\rangle$ interesting as well. It is however here interesting to note that our approach will rank the full pattern $\langle if,you,can\rangle$ higher than $\langle you,can\rangle$, because there are more compositions for patterns of length 3 than there are for patterns of length 2, and thus the mean over all those compositions is lower.}

% Algorithms~\ref{algo:ExpectedSupport} describes the computation of respectively show how to compute the set of all compositions of two sequences, the set of all binary sequential partitions of a sequence and the expected support of a sequence. 
% \begin{algorithm}
% \label{algo:Comp}
% \caption{Set of compositions}
% \KwIn{Pair of sequences $ (S_1,S_2)$}
% \KwOut{$Comp(S_1,S_2)$}
% $set\leftarrow \emptyset$ \\
% \ForEach {$\cdots$}{
% 	$set\leftarrow set \cup \cdots$
% }
% \Return $set$
% \end{algorithm}
% \begin{algorithm}
% \label{algo:BSP}
% \caption{Binary sequential partition}
% \KwIn{Sequence $S$}
% \KwOut{$BSP(S)$}
% $set\leftarrow \emptyset$ \\
% \ForEach {$\cdots$}{
% 	$set\leftarrow set \cup (\cdot,\cdot)$
% }
% \Return $set$
% \end{algorithm}
% \begin{algorithm}
% \label{algo:ExpectedSupport}
% \caption{Expected support}
% \KwIn{Sequence $S$}
% \KwOut{$ExpSupport(S)$}
% $Es\leftarrow 0$ \\
% \ForEach {$ (S_1,S_2)\in BSP(S)$}{
%    $minimum\leftarrow +\infty$\\
%    \ForEach {$S'\in Comp(S_1,S_2)$}{
%       $minimum\leftarrow \min(minimum,sup(S'))$
%    }
%    $Es\leftarrow \max(Es,minimum)$
% }
% \Return $Es$
% \end{algorithm}

\subsubsection{\rev{Our framework's null hypothesis}}
\rev{Having completely defined our proposed definition of expected support, it is important to take a step back and look at what it is achieving. \textbf{Its null hypothesis is that the frequency of a pattern can be explained by the frequency of its sub-patterns (or generators).} We have imbricated two elements: 
\begin{enumerate}
\item We define the expected frequency for each pair of sub-patterns (what we call a \emph{binary sequential partition}). Our null hypothesis there (and null model from which we get the expectancy) is that all possible ways to ``interweave'' the patterns --~such that their order is respected~-- are equally likely. The expected frequency of a pair of generators is then the average frequency observed over all possible ways to interweave these patterns. 
\item Our null hypothesis being that the frequency of a pattern can be explained by a pair of generators, the expected frequency for our framework simply considers the partition that produces the greatest expectation. 
\end{enumerate}
\paragraph{Note on the difficulty of assessing our framework's statistical significance}We will demonstrate in Section~\ref{sec:4} that our framework is very effective at extracting interesting (with all its subjectivity) patterns. It is important here to note that extracting a measure of the statistical significance of patterns under our model is a difficult task, because it requires to be able to assess the probability of a pattern given our null model. Although we can easily obtain its expected frequency, computing its probability requires more assumptions about the form of the distribution from which the data are drawn. Each sequence indeed only has a certain length, and it is unclear what model would best match our framework. We however feel that this is a consequence of the strength of our framework; we do not assume a simplistic model of the underlying distribution for all patterns. In some sense, our partitioning approach allows us to find and fit a dedicated model independently for each sequence which makes the extraction powerful, but at the cost of failing to support a simple test for assessing statistical significance. 
}

\subsubsection{Algorithmic notes}
We detail here the algorithmic considerations regarding the generation of Binary Sequential Partitions (BSPs) and of Compositions. 
\paragraph{Templates.}It is first interesting to note that in either case, the results only depend on the length of the considered pattern and are independent of the actual letters themselves. 
All partitions and compositions are thus generated as templates depending as a function of the length. 
\paragraph{Indexing templates.}Our first optimization directly follows from this observation; every time we create a template for a BSP  of a particular length, we index it with its length so that it is only computed once for each length. 
We employ a similar indexing for Compositions, with the difference that they have two associated lengths; we thus index them using a matrix. 
Note that the template for a Composition of length $l_1$ and $l_2$ is identical, we only use the upper triangle in the matrix. 
\paragraph{Generating BSPs.}Generating all Binary Sequential Partitions for a template of length $l$ is an enumeration exercise. 
The first element to note is that knowing the pattern and its left partition set ($S_1$ in Definition~7) completely determines its right partition set ($S_2$ in Definition~7), and conversely. 
This means that, to construct the template for all BSPs of a particular length, we only need to find what is the left partition set, \ie{}the sequence of positions of elements in the original pattern. 
Furthermore, for a pattern of length $l$, we only need to consider $S_1$ of lengths $l_1$ from 1 to $\lfloor\frac{l}{2}\rfloor$, because of symmetry.

\rev{To generate all possible templates for $S_1$, we enumerate all the $l_1$-combinations of the set of $l$ positions in $S$; there are ${l \choose l_1}$ such combinations. }
To this end, we use the standard enumeration algorithm for combinations as described in \cite{Tucker2006}. 
\paragraph{Generating Combinations.}Generating combinations is simpler than BSPs. The same symmetry observations hold here, \ie{}the template for all combinations of lengths $ (l_1,l_2)$ is the same as the one for all combinations of lengths $ (l_2,l_1)$. 
\rev{We then generate all the possible $l_1$-combinations of the set of $l_1+l2$ positions;} here again we use the standard algorithm described in \cite{Tucker2006}.

\subsection{Exploring the space of all sequential patterns}
\label{sec:3.3}
%At first consideration, search for such sequential patterns appears computationally intractable. The average search space of potential sequences is $L^{|\mathcal{I}|}$, where $L$ is the average length of sequence in $\mathcal{D}$. It is clear that efficient search of this space will require very efficient pruning mechanisms. 
Our algorithm to explore the space of sequential patterns and exactly extract the top-$k$ sequential patterns with highest leverage\footnote{In the remainder of this paper, we will use the term 'leverage' as a proxy for 'leverage under our definition of expected support'. } is based on the OPUS Miner algorithm \cite{RefWebb1995,RefWebb2014}, first introduced for mining non-sequential databases. 

\subsubsection{Sequential top-$k$ OPUS Miner (SkOPUS)}
OPUS Miner finds the top-$k$ itemsets under given interestingness measures from a non-sequential database, with regard to a given measure of interest.
OPUS Miner is a depth-first branch-and-bound algorithm, that exploits the monotonicity of itemsets ($sup(I\cup i)\leqslant sup(I)$). \rev{It performs an exhaustive depth-first search of the space of sequential patterns except in so far as it can prune parts of the search space that cannot contain itemsets in the top-$k$ with respect to the measure of interest.} For example, if the minimum (\textit{i.e.} worse) value in the top-$k$ is $\alpha$, and we know that any specialization of an itemset $I$ will have an score that is lower than $\alpha$, then the sub-tree for which the root is $I$ does not need to be explored. 

OPUS Miner's depth-first search has the advantage that the number of open nodes at any one time is minimized. This allows extensive data to be maintained for every open node and ensures that only one such record will be stored at any given time of each level of depth that is explored. It also promotes locality of memory access, as most new nodes that are opened are minor variants of the most recently explored node. %It has the disadvantage that larger numbers of deep nodes are explored than would be necessary if breadth-first search were employed. This presents a significant computational burden due to the complexity of node evaluation increasing exponentially as itemset size increases; Section~\ref{subsubsec:bootstrap} will show this burden can be minimised.
%To extract the top-$k$ sequential patterns with highest leverage, we have adapted the way OPUS explores the search space to account for the specificity of sequential databases.%, and to ensure that the top-$k$ sequential patterns are exactly the same ones as the ones that would have been extracted if the complete search was explored. Moreover, the order in which the space is explored has to make sure that the pruned sub-spaces are not re-consider later in the search. 

Algorithms~\ref{algo:SkOPUS} and \ref{algo:ExpandSequence} describe SkOPUS.%, while an illustration of the way SkOPUS explores the search is given in Figure~\ref{fig:search}. 
The main elements that differ from the original OPUS Miner algorithm are as follows: 
\begin{enumerate}
\item When a sequential pattern is specialized, the specializing item is always placed at the end of the sequential pattern (as a suffix~--~appending item $i$ to sequence $S$ is noted $S.i$ hereafter). This ensures that sub-spaces that are pruned will not be reconsidered later in the search. 
\item When retrieving an item from the queue of available items to specialize a pattern with, the item is left in the queue (as opposed to OPUS Miner where the item is removed from the queue). This is necessary to handle repetitions of items in the patterns such as $\langle a, a, b\rangle$. \rev{Thus, we use the variant of OPUS \cite{RefWebb1995} that allows multiple applications of a single operator (the addition of a given item).}
\item Upper bounds for any sequential extension of a pattern $S$ have to be adapted for our measure of interest (leverage) and our definition of expected support; we detail this element in Section~\ref{subsubsec:bounds}. 
\end{enumerate}
\begin{algorithm}
\label{algo:SkOPUS}
\caption{SkOPUS Miner}
\KwIn{Sequential database $\mathcal{D}$, a measure of interest $M$, integer $k$}
\KwOut{The top-$k$ sequential patterns with regard to $M$}
$q\leftarrow$ a queue of all items $i\in\mathcal{D}$ in descending order on $\supp(\langle i\rangle)$\;
$topK\leftarrow$ an empty queue to contain the top-$k$ sequential patterns\;
\Return ExpandSequence($\langle \rangle$, $q$, $topK$, $M$)\;
% \For {all $i\in{q}$}{
%     $S\leftarrow \langle i\rangle$\\
%     ExpandSequence($S$, $q$, $topK$, $M$)
% }
\end{algorithm}
\begin{algorithm}
\label{algo:ExpandSequence}
\caption{ExpandSequence}
\SetKwData{Index}{Index} 
\KwIn{A sequence $S$, a queue $q$, the current $topK$ and a measure of interest $M$}
\KwOut{The top-$k$ sequential patterns with regard to $M$ in the search space explored to date}
%\KwOut{the most $K$ interesting sequence}
initialize $q'$ to be an empty queue of items\;
\For {all $i\in{q}$}{
	$T \leftarrow \brak{S \mid i}$\;
	%$Es \leftarrow \esupp(T)$\;
    $score \leftarrow M(T)$\;
   \If {$score > topK.min$}{
      $topK.add(T,score)$\;
   }
   \If {$Upper\_bound(M,T) > topK.min$}{
      $q'\leftarrow append(q', i)$\;
   }
}
\For {all $i\in q'$ in descending order w.r.t. $M$}{
   $topK\leftarrow $ExpandSequence($T$, $q'$, $topK$, $M$)\;
}
\Return $topK$\;
\end{algorithm}
% \begin{figure}
% 	\centering
%     \includegraphics[width=0.75\textwidth]{fig1searchspace}
%     \caption{\label{fig:search}Example of search space exploration for a sequential database containing 3 possible elements $a$, $b$ and $c$.\fpcomment{not needed, quite standard}}
% \end{figure}
\rev{Algorithm~\ref{algo:SkOPUS} simply initializes the necessary data structures to then call Algorithm~\ref{algo:ExpandSequence} that systematically explores the entire space of sequential patterns other than those branches of the search space of which it can be determined no top-$k$ can be contained therein. The correctness of SkOPUS follows from the correctness of OPUS. As SkOPUS investigates each pattern it maintains a list $topK$ of the top-$k$ patterns found so far. Thus, on termination, as the entire search space has been explored other than those branches that cannot contain patterns in the top-$k$, $topK$ contains the top-$k$ patterns.}

\subsubsection{Upper bounding}
\label{subsubsec:bounds}
We now detail how to upper bound the score of any extension of a sequential pattern $S$ with regard to leverage (\ie{}how to compute $Upper\_bound(M,S\star)$ in Algorithm~\ref{algo:ExpandSequence}). As presented earlier, we measure the interestingness of a pattern with a \emph{leverage}, 
\[
	\leverage = \supp(S)-\esupp(S).
\]
In order to prune the search space we need an upper-bound on $\leverage$ for any extension pattern, say $T$, of a pattern $S$.
Since $\esupp(S) \geq 0$  and $\supp(T) \leq \supp(S)$, we immediately obtain $\leverage(T) \leq \supp(S)$, allowing us to use $\supp(S)$ as an upper bound. \rev{Note that, possibly surprisingly, the lower bound on $\esupp(S)$ is tight. Consider the example of a database where all sequences start with the same pattern $P$ of length $\ell$, and then do not use any of the elements composing the pattern in the rest of the sequence. We will then have all sequential compositions of all possible binary sequential partitions with 0 frequency, except for the one with the actual pattern. The highest frequency will then be for the one with the lowest number of compositions, which is $\ell$. 
\begin{displaymath}
\esupp(P)\geqslant\frac{\supp(P)}{\ell}
\end{displaymath}
Given that we want to bound all possible extensions of $P$, $\ell$ is unbounded, which makes $\esupp(P\star)>0$.}

\iffalse
\fpcomment{too verbose}
To upper bound any ``suffixed extension'' of a pattern $S$ (adding an item as a suffix of $S$), noted $S\star$, we need to:
\begin{enumerate}
\item upper bound $sup(S\star)$
\item lower bound $ExpSupport(S\star)$
\end{enumerate}
We use the following properties: 
\begin{eqnarray}
sup(S\star)&\leqslant& sup(S)\\
ExpSupport(S\star)&\geqslant& 0
\end{eqnarray}
This leads directly to the following upper bound for $leverage$:
\begin{equation}
leverage(S\star) \leqslant sup(S)
\end{equation}
\fi
% Note that there is no upper-bound for the lift if we use the maximum likelihood estimates for the support. This is why this paper focuses mainly on the use of \emph{leverage}, which constitutes a good measure for our framework. We nonetheless present some first results using the \emph{lift} in our experiments on literary work, which takes advantage of $m$-estimates for the support. We can then derive the following upper bound for $lift$:
% \begin{equation}
% lift(S\star) \leqslant \frac{sup(S)}{ExpSupport(S\star)} \leqslant \frac{\#(S)+m}{m}
% \end{equation}
% Where $m=1$ is a standard value and $\#(S)$ is the size of the cover of $S$. 

%\subsubsection{SkOPUS can be used for any definition of expected support}
In the definitions above, we exploit the fact that support and expected support are positive. It is interesting to note that, so long as the definition of expected support does not allow negative values, these elements will remain valid. It follows that our SkOPUS algorithm can be directly used with any sensible definition of expected support.

\subsubsection{Bootstrap --- How to quickly get a good top-$k$?}
\label{subsubsec:bootstrap}
%The worst-case time complexity of our algorithm with respect to the number of items is $\Omega(|I|^{maxL})$, where $maxL$ is the maximum length of the sequence in dataset. 
In practice, the time requirements of SkOPUS depend on the efficiency of the pruning mechanisms, and can vary greatly from dataset to dataset\footnote{Note that if, for instance, most of the sequential patterns in a dataset are non-interesting, then the algorithm will have to explore a large part of the space.}.  SkOPUS traverses the search space maintaining the set of top-$k$ sequential patterns discovered in the search space explored so far. \rev{The search space can only be pruned of branches that cannot contain a pattern with higher leverage than the $k^{th}$ best leverage found so far.} Thus, efficient pruning relies critically on our ability to quickly fill $topK$ with high-scoring patterns so that as much of the search space can be pruned as possible. 

We use two main strategies. First, we consider the addition (as suffix) of items with highest support first; this is apparent in the order in which we go through the different queues in Algorithms~\ref{algo:SkOPUS} and \ref{algo:ExpandSequence}.

\rev{Algorithm~\ref{algo:ExpandSequence} performs a depth-first search. The primary reason for doing so is that it limits the number of nodes in the search space that must be simultaneously open. To efficiently compute the cover of a new node it is important to store the cover of the parent and only update it with respect to the item appended to the sequence. Minimizing the number of open nodes minimizes the number of covers that must be stored. However, we store the cover of every item, and so this issue does not affect two-item sequences. Our second strategy uses a hybrid search. First a breadth-first search is performed over all two-item sequences. Then the regular depth-first search is employed for sequences of length three and greater. We present the two-item breadth-first bootstrap in Algorithm~\ref{algo:bootstrap}. The algorithm remains correct as it still systematically searches all of the search space that might contain top-$k$ patterns, all that changes is the order in which they are explored.}
%Second, we propose a bootstrap mechanism that performs a quick breadth-first search with limited depth; Algorithm~\ref{algo:bootstrap} illustrates our bootstrap breadth-first mechanism with a maximum depth of 2. Note that the maximum depth of the bootstrap can be easily adapted depending on the domain and size of vocabulary. 
\begin{algorithm}
\label{algo:bootstrap}
\caption{SkOPUS Bootstrapping}
\KwIn{Sequential database $\mathcal{D}$ and a measure of interest $M$}
\KwOut{A prefilled $topK$}
Let $topK$ be an empty ordered queue\\
\ForEach{$i_1, i_2 \in\mathcal{I}$}{
	$S \leftarrow \langle i_1,i_2\rangle$\\
	$Es \leftarrow \esupp(S)$\\
	$score \leftarrow M(S, Es)$\\
   \If {$score > topK.min$}{
	  $topK \leftarrow topK.add(S,score)$
   }
}
\Return $topK$
\end{algorithm}

\section{Experiments}
\label{sec:4}
In this section, we present the results of SkOPUS for the exact extraction of the $k$ sequential patterns with highest leverage. In the synthetic experiments we compare the performance of SkOPUS using leverage as the measure of interest against SkOPUS using support as the measure of interest (included to provide a contrast against frequent pattern discovery), state-of-the-art sequential pattern discovery algorithm $\NM$ \cite{Tatti2015-partition} and baseline $\Mind$ \cite{Gwadera2005-independence}.  As the synthetic experiments clearly show the baseline is  less effective than either SkOPUS using leverage or $\NM$, we do not include it in the real-world experiments.

\subsection{Datasets}
We shall start this section by making a general observation about the availability of interpretable sequential databases. 
Applications of sequential pattern mining methods are numerous and include: 
\begin{enumerate}
\item sequences of the different webpages browsed by users on a website, where each visit represent a new transaction in the database;
\item locations of series of events, such as the neighborhoods that a taxi drives through for each client, or the restaurants visited by registered customers of say Yelp\textregistered{} or Urbanspoon\textregistered{};
\item the performance of different industries on the share market over different days of trading; 
\item the sequence of actions performed by a surgeon over the course of different surgeries; 
\item the tests performed on (and action taken about) patients from admission to discharge. 
\end{enumerate}
The datasets associated to all these applications are however extremely valuable and hence only rarely made available freely to the scientific community. 
In addition, assessing the quality of technologies for pattern discovery requires having knowledge about the patterns that are actually present in data. 
Therefore we start by evaluating the discovery with sequences that are sampled from known distributions with specific patterns. We can then compare the discovered interactions to the true structure from which the data was sampled. This is the first set of experiments we present in Section~\ref{exp:synthetic}. 
%Note that our aim is, upon acceptance of the paper, to provide the first data archive to assess sequential pattern mining technologies. 
Then, in Section~\ref{exp:real}, we assess the relevance and scalability of our approach on public domain literary works. We have selected several books which we believe have characteristics that are quite representative of many other types datasets. 

It is important to note that the main objective of this paper is not to prove that extracting patterns from sequential databases is an important topic; this has in fact been largely motivated by the data mining community. 
Rather, we aim at demonstrating 1. that interestingness-based sequential pattern extraction is critical and 2. the relevance of our approach.
%; although we would have preferred to present more exhaustive results, we believe that these experiments sufficiently support both of these claims. 

%Note that we are in the process of establishing the first sequential pattern mining data archive.
%We will make datasets publicly available with their associated ``ground truth'', \textit{i.e.} the set of patterns embedded in the sequences with their associated probability. 
All the datasets used in this paper (as well as their ``ground truths'', when available) are provided for reviewing purposes at \url{https://github.com/fpetitjean/Skopus}.

\subsection{Experiments on data with known patterns}
\label{exp:synthetic}
We first assess our method on datasets that embed a known set of sequential patterns. 

Details of the data generation process are as follows: we start by generating a dataset with 10,000 random sequences over a vocabulary $\mathcal{I}$ with 10 tokens; the probability of the tokens follows a flat Dirichlet distribution: $\mathcal{I}\sim\operatorname{Dir}(\vec{1})$. We generate each random sequence in the dataset by choosing its length $l_s$ distributed accordingly to a shifted\footnote{Shifting ensures that the sequences have at least 2 elements. } Poisson distribution: $l_s\sim \operatorname{Pois}(9.0)+1$. This makes the sequences to have an average length of 10 with standard deviation 3 ($E(l_p)=10$, $\operatorname{Var}(l_p)=9$).

We then generate a set of $k$ sequential patterns, each with an associated probability of occurrence in a sequence drawn uniformly in $[0.05,0.2]$ and length $l_p$ following $l_p\sim \operatorname{Pois}(1.0)+2$, \textit{i.e.} with average length 3 and standard deviation 1 ($E(l_p)=10$, $\operatorname{Var}(l_p)=1$).  
Then, each pattern is embedded sequentially according to its associated probability. Embedding is performed by uniformly at random selecting insertion points in the sequence. 

We then extracted the top-20 sequential patterns according to $leverage$ and compare the results with a support-based extraction\footnote{Note that we ``help'' the support-based extraction by not allowing it to test single items, otherwise it would only return those.}, $\NM$~\cite{Tatti2015-partition} and $\Mind$, a baseline used by~\cite{Tatti2015-partition} using the indepdence model
proposed by~\cite{Gwadera2005-independence}. Both baselines require a set of candidate patterns: here we used frequent patterns with a threshold of 500. Note that this threshold potentially ``helps'' these two methods, because none of the injected patterns have a support lower than 500; such a threshold has thus potential to prune a significant number of patterns that could have been ranked in the top-10. 
%which we know are not patterns. 
\begin{table}
\centering
\caption{\label{tab:res-synthetic}Results on data with known embedded patterns}
\begin{tabular}{lrrrr}
\toprule
\#patterns&  \multicolumn{4}{l}{Recall}\\[2pt]
\cmidrule{2-5}
  & Support  & Leverage & $\NM$ \cite{Tatti2015-partition}& $\Mind$ \cite{Gwadera2005-independence}\\
%$|\mathcal{D}|$ & \#patterns& \# found & average rank & \# found & max $k$& avg $k$\\
\midrule
 1  & 0\% & \textbf{100\%} & \textbf{100\%}& \textbf{100\%} \\[2pt]
 2  & 0\% & \textbf{100\%} & \textbf{100\%}& 50\%\\[2pt]
 3  & 0\% & \textbf{100\%} &66\% & 66\%\\[2pt]
 4  & 25\% & \textbf{100\%} & \textbf{100\%} & 75\%\\[2pt]
 5  & 0\% & \textbf{100\%} & 40\% & 40\%\\[2pt]
 6  & 33\% & \textbf{67\%} & \textbf{67\%} & 33\%\\[2pt]
 7  & 14\% & \textbf{86\%}& 43\% & 43\% \\[2pt]
 8  & 0\% & \textbf{25\%}& 12\% & 12\%\\[2pt]
 9  & 0\% & \textbf{78\%}& 44\% & 22\% \\[2pt]
10  & 0\% & \textbf{40\%}& 20\% & 10\% \\[2pt]
\bottomrule
\end{tabular}
\end{table}
We report recall rates in Table~\ref{tab:res-synthetic}, that is, proportions of all patterns that were embedded that are included in the top-20 patterns returned by each approach.  As expected, the support-based method does not perform well. In fact, it only ever extracted patterns of length 2; which mainly correspond to sequential patterns that are frequent because their composing items are as well. This is a similar behaviour to the one that we have explained with \emph{and}, \emph{to} and \emph{of} in the introduction: patterns composed with frequent items will appear frequently by chance, without being interesting. This means that top-$k$ extraction based on support has difficulty extracting patterns of length $\geqslant 3$. 

To further highlight this point, we shall mention that for the experiment with a single embedded pattern ($\#\text{patterns}=1$ in Table~\ref{tab:res-synthetic}),
$\langle c, b, c, a\rangle$ embedded in 19\% of the sequences, this pattern is ranked 92\textsuperscript{nd} under support while it is the top pattern under our $\leverage$, $\NM$ and $\Mind$.
More generally, Table~\ref{tab:res-synthetic} shows that all approaches outperform support-based approaches, by recovering a much larger number of embedded patterns. 

We can also observe that our method ---~SkOPUS with leverage~--- outperforms $\NM$ and $\Mind$  by obtaining a significantly higher recall of the embedded patterns. It is also interesting to note that these experiments confirm the ones in \cite{Tatti2015-partition} by showing the poorer performance of $\Mind$ compared to $\NM$. 

It can be observed that, as the number of patterns embedded in the data increases, the recall decreases. This is due to three main factors: 
\begin{enumerate}
\item because we keep the number of patterns extracted constant ($k=20$), it is normal that  recall diminishes as the number of actual patterns increases;
\item our approach considers the subsequences of an embedded pattern to also be interesting patterns: this means even if we were to extract an actual pattern $\langle d,m,k,d\rangle$, we are also prone to extract subsequences of this patterns, such as $\langle d,m,k\rangle$ or $\langle d,m,d\rangle$. There is no counterpart of \emph{independent productivity} from self sufficient itemset mining whereby subpatterns can be discarded \cite{RefWebb2010}.
\item some patterns can overlap; for example if we have two (actual) patterns $\langle a,e,e,i\rangle$ and $\langle e,f,i,d\rangle$ that are independently embedded in the data, then, the pattern $\langle e,i\rangle$ is going to be even more frequent and can be extracted; even though it is not one of the directly embedded patterns. 
\end{enumerate}

Note that the two last points actually complicate the extraction of patterns within the top-20 as well. These three elements are best exemplified with the top-20 sequential patterns corresponding to the dataset containing 7 patterns, which we illustrate in Table~\ref{table:data-7} and Table~\ref{table:res-synthetic-2}, where we report the exact matches and subpatterns that are discovered, adopting the elements appropriate to sequential patterns from the approach pioneered by Zimmermann \cite{RefZimmermann2013} for non-sequential patterns. 
\begin{table}
\centering
\rev{
\caption{\label{table:data-7}Details on the datasets embedding 7 patterns. }
\begin{tabular}{lll}
\toprule
Injected patterns  &Support&Leverage\\[1pt]
\midrule
$\langle i, g, d\rangle$&17.1\% &3.36\% \\[1pt]
$\langle d, e\rangle$ &15.9\% & 0.85\%\\[1pt]
$\langle a, e, j\rangle$& 14.6\% & 2.95\%\\[1pt]
$\langle i, b\rangle$ &14.2\% & 4.26\%\\[1pt]
$\langle j, e, j, g\rangle$ & 10.5\% & 2.11\%\\[1pt]
$\langle j, d, c, i, a\rangle$ &7.8\% & 3.58\%\\[1pt]
$\langle j, h, i\rangle$ & 7.1\% & 2.34\%\\[1pt]
\bottomrule
\end{tabular}}
\end{table}

\begin{table}
\centering
\caption{\label{table:res-synthetic-2}Top 20 results on the dataset embedding 7 patterns. Exact matches are represented in blue with a $\bullet$ next to it; subsequences of such exact matches are depicted in boldface.}
\begin{tabular}{lrrrr}
\toprule
\#&Top-$k$\\[2pt]
\cmidrule{2-5}
 & Leverage (value)& Support& $\NM$ \cite{Tatti2015-partition}& $\Mind$ \cite{Gwadera2005-independence} \\[1pt]
\midrule
1&$\mathbf{\langle i, g\rangle}$ (4.3\%) &$\mathbf{\langle j, d\rangle}$&\textcolor{blue}{$\mathbf{\bullet\langle j, d, c, i, a\rangle}$} &\textcolor{blue}{$\mathbf{\bullet\langle j, d, c, i, a\rangle}$}\\[1pt]
2& \textcolor{blue}{$\mathbf{\bullet\langle i, b\rangle}$} (4.3\%)&$\langle j, e\rangle$&\textcolor{blue}{$\mathbf{\bullet\langle i, b\rangle}$}&$\langle d, d, c, i, a\rangle$\\[1pt]
3& \textcolor{blue}{$\mathbf{\bullet\langle j, d, c, i, a\rangle}$} (3.6\%) &$\langle d, d\rangle$&\textcolor{blue}{$\mathbf{\bullet\langle i, g, d\rangle}$}&$\langle j, j, c, i, a\rangle$\\[1pt]
4&\textcolor{blue}{$\mathbf{\bullet\langle i, g, d\rangle}$} (3.4\%) &$\mathbf{\langle j, j\rangle}$&$\mathbf{\langle i, g\rangle}$ &$\mathbf{\langle j, c, i, a\rangle}$\\[1pt]
5 & $\mathbf{\langle d, c, i, a\rangle}$ (3.2\%)&\textcolor{blue}{$\mathbf{\bullet\langle d, e\rangle}$}&$\mathbf{\langle j, c, i, a\rangle}$&$\mathbf{\langle d, c, i, a\rangle}$ \\[1pt]
6 &$\mathbf{\langle j, c, i, a\rangle}$ (3.2\%) &$\langle d, j\rangle$& $\langle j, j, c, i, a\rangle$&$\langle j, e, c, i, a\rangle$\\[1pt]
7 & \textcolor{blue}{$\mathbf{\bullet\langle a, e, j\rangle}$} (2.9\%)&$\mathbf{\langle e, j\rangle}$&$\langle d, d, c, i, a\rangle$&$\langle j, e, e, j, g\rangle$\\[1pt]
8 &$\langle i, g, g\rangle$ (2.8\%) &$\langle i, d\rangle$&$\langle i, g, g\rangle$&$\langle d, j, c, i, a\rangle$\\[1pt]
9&$\langle i, e\rangle$ (2.7\%) &$\langle e, e\rangle$& $\mathbf{\langle d, c, i, a\rangle}$&$\langle j, d, c, e, a\rangle$\\[1pt]
10&$\mathbf{\langle j, d, c, a\rangle}$ (2.6\%)&$\mathbf{\langle j, g\rangle}$ &$\langle i, g, g, d\rangle$&$\langle i, i, b\rangle$\\[1pt]
11 &$\mathbf{\langle j, d, c, i\rangle}$ (2.6\%) &$\mathbf{\langle i, d\rangle}$ &$\langle j, e, c, i, a\rangle$&$\mathbf{\langle j, d, c, a\rangle}$\\[1pt]
12&\textcolor{blue}{$\mathbf{\bullet\langle j, h, i\rangle}$} (2.3\%) &$\mathbf{\langle g, d\rangle}$ &$\langle j, h, i\rangle$&\textcolor{blue}{$\mathbf{\bullet\langle i, g, d\rangle}$}\\[1pt]
13 &$\langle a, e, e, j\rangle$ (2.3\%)&$\langle d, g\rangle$ &$\mathbf{\langle j, d, c, a\rangle}$&$\mathbf{\langle j, d, c, i\rangle}$\\[1pt]
14 &$\langle a, e, e\rangle$ (2.3\%)  &$\langle i, j\rangle$&$\langle d, j, c, i, a\rangle$& \textcolor{blue}{$\mathbf{\bullet\langle j, e, j, g\rangle}$}\\[1pt]
15 &$\mathbf{\langle j, d, i, a\rangle}$ (2.2\%)&$\langle i, e\rangle$ &$\langle j, j, e, j, g\rangle$&$\langle i, g, g, d\rangle$\\[1pt]
16 & \textcolor{blue}{$\mathbf{\bullet\langle j, e, j, g\rangle}$} (2.1\%)&$\mathbf{\langle e, g\rangle}$ &$\langle j, e, e, j, g\rangle$&$\langle j, d, e, j, g\rangle$\\[1pt]
17 &$\mathbf{\langle j, e, g\rangle}$ (2.1\%) &$\langle g, j\rangle$&$\langle i, e, d\rangle$&$\langle j, e, j, j, g\rangle$\\[1pt]
18 &$\langle a, e, g\rangle$ (2.1\%) &$\mathbf{\langle j, i\rangle}$&$\langle i, g, d, d\rangle$&$\langle j, d, c, i, g\rangle$\\[1pt]
19 &$\langle i, g, g, d\rangle$ (2.1\%)&$\mathbf{\langle d, i\rangle}$&$\langle i, c, i, a\rangle$&$\langle j, d, c, j, a\rangle$\\[1pt]
20 &$\langle i, e, e\rangle$ (2.0\%) &$\langle g, e\rangle$&$\langle j, d, c, e, a\rangle$&$\langle j, d, d, i, a\rangle$\\[1pt]
\bottomrule
\end{tabular}
\end{table}
First, this table further illustrates the consistency of our method. We have depicted the extracted patterns that are actual injected patterns in with a $\bullet$ next to them. While our method can extract 6 out the 7 actual patterns within the top-$20$, we can see that the support-based method only extracts one pattern (corresponding to a pair), and the two other methods only 3 actual patterns. 
It is also interesting to note that the similarity between the 4 first patterns extracted by our method and by $\NM$.

\rev{It is informative to explain why the pattern $\langle d, e\rangle$ is not part of our top-20 while it is part of the top-20 on the $support$: it it is quite a frequent pattern (support of 63\%), but this is actually mostly due to the support of $d$ and $e$. This pattern was introduced with a 16\% probability; the difference is explained by the fact that both tokens $d$ and $e$ were sampled with a relatively high probability $\geqslant$ 15\%. }This means that the support method only extracts it in the top-20 because $d$ and $e$ are frequent to start with; had they been infrequent tokens, the support wouldn't have ranked the pattern that high. This is exactly what happens with pattern $\langle i, b\rangle$, for which token $b$ has much lower probability ($p(\langle b\rangle)=0.02$). Because it does not appear \textbf{by chance} frequently, the support-based approach ranks the pattern very low. It is also interesting to see that under $support$, pattern $\langle e, d\rangle$ is only slightly less interesting, which clearly shows its inconsistency.  
Conversely, we can see the consistent behaviour of the leverage-based approach, which ranks $\langle i, b\rangle$ with much higher value than $\langle d, e\rangle$. Before this pattern was embedded into the data it already appeared in about 61\% of the sequences. Observing it in 63\% is thus not a very strong effect and our method ranks it accordingly. On the other hand, before introduction of $\langle i, b\rangle$, the pattern only appeared in about 13\% of the sequences, observing it 22\% is thus more significant and our method correctly ranks it accordingly.

Moreover, we have also depicted in boldface (with no $\bullet$) the patterns extracted by our approach that correspond to subsequences of actual patterns that we have also extracted. This mainly illustrates the two first elements that we have noted, by having 8 slots in our top-20  ``consumed'' by sub-patterns of high leverage. 
Although this falls out of the scope of this first attempt at extracting the top-$k$ sequential patterns with leverage, this is naturally echoing the work that has been done on filtered-top-$k$ association discovery \cite{RefWebb2010,RefWebb2011}. 
More generally, these results call for a reflection on the evaluation of sequential pattern mining procedures, similarly to the work that has been performed in this area for non-sequential pattern mining \cite{RefZimmermann2013}. 

\subsection{Experiments on literary works}
\label{exp:real}
We compared the top patterns from several works of literature  and the JMLR abstract dataset. For brevity, we only discuss the patterns obtained from JMLR. For more information about other results, see our section Supplementary material at the end of the manuscript. 

We study in detail the results of the different methods on the JMLR dataset \cite{Tatti2012}, which represents the abstracts of the papers published in the \emph{Journal of Machine Learning Research}. This dataset holds 788 abstracts (hence sequences), which use $3,844$ words (items). The average length of abstracts is 96 words with a maximum length of 231. Since $\NM$ requires a candidate set, we use frequent patterns with a threshold of 10. For the other literature works we used a threshold of 5.

We first present the results and detail the computation times in the next subsection. 
% We first present the results obtained considering one sequence per sentence, then one sequence per paragraph, and then detail the computation times. 
The top-$10$ patterns are presented in Table~\ref{table:res-JMLR} for our method (leverage), support for reference, and the results from $\NM$; having shown in the previous section that our method and $\NM$ outperform $\Mind$, we focus on other methods and keep top-support patterns for reference.

\begin{table}
\centering
\caption{\label{table:res-JMLR}Top 10 results on abstracts of JMLR papers}
\begin{tabular}{llll}
\toprule
&  \multicolumn{2}{l}{Top-$k$ patterns}\\[2pt]
\cmidrule{2-4}
\#&Support & Leverage & $\NM$ \cite{Tatti2015-partition}  \\[1pt]
\midrule
1&$\langle$\emph{algorithm, algorithm}$\rangle$  &  $\langle$\emph{paper, show}$\rangle$  & $\langle$\emph{support, vector}$\rangle$\\[1pt]%& $\langle$\emph{support, vector, machin}$\rangle$  \\[1pt]
2&$\langle$\emph{learn, learn}$\rangle$  &  $\langle$\emph{paper, result}$\rangle$  & $\langle$\emph{support, vector, machin}$\rangle$\\[1pt]% & $\langle$\emph{support, vector, machin, svm, svm}$\rangle$ \\[1pt]
3&$\langle$\emph{learn, algorithm}$\rangle$   & $\langle$\emph{support, vector, machin}$\rangle$  & $\langle$\emph{support,  machin}$\rangle$ \\[1pt]%& $\langle$\emph{support, vector, machin, svm, svm, svm}$\rangle$ \\[1pt]
4&$\langle$\emph{algorithm, learn}$\rangle$   &  $\langle$\emph{paper, algorithm}$\rangle$ & $\langle$\emph{real, world}$\rangle$  \\[1pt]%& $\langle$\emph{support, vector, machin, svm}$\rangle$\\[1pt]
5&$\langle$\emph{data, data}$\rangle$  & $\langle$\emph{support, vector}$\rangle$    &   $\langle$\emph{vector, machin}$\rangle$   \\[1pt]%& $\langle$\emph{support, vector}$\rangle$\\[1pt]
6&$\langle$\emph{learn,data}$\rangle$   &   $\langle$\emph{base, result}$\rangle$ &  $\langle$\emph{state, art}$\rangle$      \\[1pt]%& $\langle$\emph{support, vector, machin, svm, svm, svm, svm}$\rangle$\\[1pt]
7&$\langle$\emph{model, model}$\rangle$   &  $\langle$\emph{paper, method}$\rangle$  &  $\langle$\emph{reproduc, hilbert}$\rangle$  \\[1pt]%& $\langle$\emph{reproduc, kernel, hilbert, space}$\rangle$ \\[1pt]
8&$\langle$\emph{problem, problem}$\rangle$    &  $\langle$\emph{learn,result}$\rangle$  &  $\langle$\emph{high, dimension}$\rangle$ \\[1pt]% & $\langle$\emph{support, vector, machin, support, vector}$\rangle$  \\[1pt]
9&$\langle$\emph{learn, result}$\rangle$   &  $\langle$\emph{paper, propos}$\rangle$  &  $\langle$\emph{first, second}$\rangle$    \\[1pt]%& $\langle$\emph{kernel, kernel, kernel, kernel, kernel, kernel, kernel}$\rangle$\\[1pt]
10&$\langle$\emph{problem, algorithm}$\rangle$    &  $\langle$\emph{vector,machin}$\rangle$  &  $\langle$\emph{larg, scale}$\rangle$\\[1pt]%& $\langle$\emph{support, vector, machin, result}$\rangle$  \\[1pt]
\bottomrule
\end{tabular}
\end{table}

\begin{figure}
\centering
\subfigure[\label{subfig:overlap-100}Top-100]{\includegraphics[width=.46\linewidth]{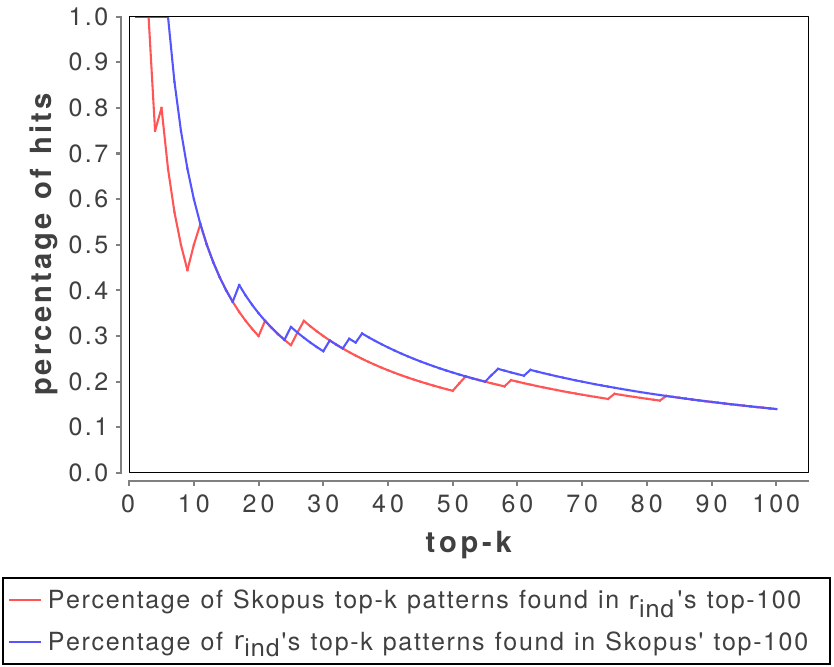}}
\subfigure[\label{subfig:overlap-1000} Top-1000]{\includegraphics[width=.46\linewidth]{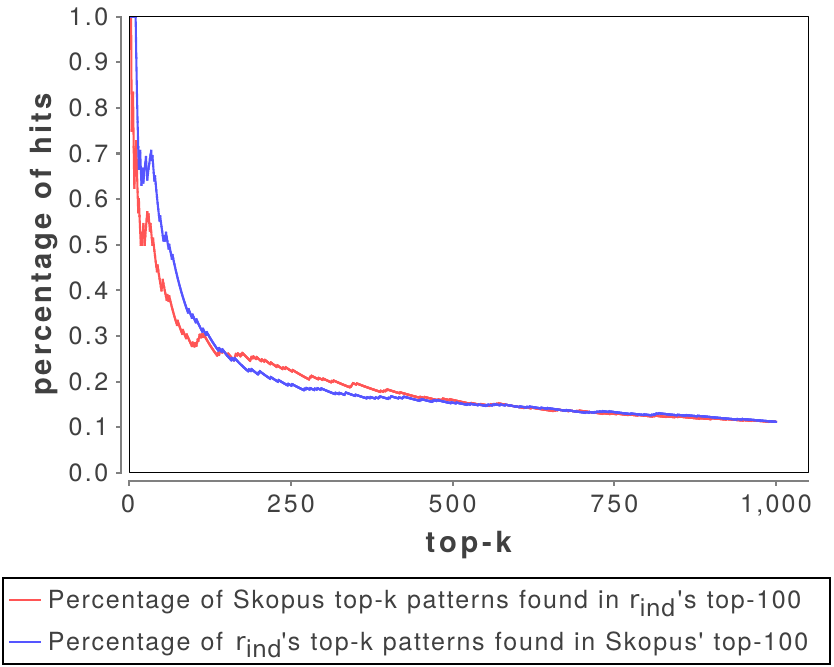}}
\caption{\label{fig:overlap}Overlap between the top-$k$ given by SkOPUS and $\NM$}
\end{figure}

The first critical observation echoes the ones we made in the introduction about the inconsistency of support-based extraction: when using the support, most of the patterns correspond in repetitions of very frequent words, such as $algorithm$, $learn$, $data$, $model$ and $problem$. Moreover we can see that when using the support as the measure of interest, the pattern $\langle learn, algorithm\rangle$ and its reversed version $\langle algorithm, learn\rangle$ appear with very similar scores (resp. supports 36\% and 29\%). 
In contrast, we can observe that our method and $\NM$ present patterns that seem of higher general interest than support. 
Interestingly, the patterns extracted seem to differ significantly between the methods. 
Beyond highlighting the importance of synthetic data experiments, this also re-confirms the subjectivity of interestingness. We find then interesting to examine two elements: 
\begin{enumerate}
\item The overlap of the different methods.
\item The patterns that crystallize the differences between the methods. 
\end{enumerate}
\paragraph{Overlap between SkOPUS and $\NM$}
For each method, we examine what percentage of its top-$k$ is found in the top-$k$ of the other.
Figure~\ref{fig:overlap} presents this overlap analysis for SkOPUS and $\NM$: in \subref{subfig:overlap-100} their top-100 and \subref{subfig:overlap-1000} their top-100. For example, a point at coordinates $ (x,y)$ for the top-100 tells us that $y$ percent of the top-$x$ of the first method was found in the top-100 of the other. 
Two similar results would then have a slowly decreasing rate. 
In this case we observe quite the opposite: there is only little overlap between the two methods with less than 20\% overlap within the top-100 and top-1000. 
Interestingly, SkOPUS seem to find more of the top patterns ranked by $\NM$ than the opposite, which supports the relevance of our method. 

\paragraph{Analysis of representative differences}
Significant differences exist even in the very top patterns, which clearly appears in Figure~\ref{subfig:overlap-100}, which shows that patterns as early as in SkOPUS' third position are not part of the top-100 of the patterns ranked by $\NM$. 
We detail below a few such example patterns that are highly contrasting the differences between SkOPUS and $\NM$: 
\begin{itemize}
\item Patterns SkOPUS ranks high but not $\NM$: $\langle paper, algorithm\rangle$ is extracted as the  4\textsuperscript{th} pattern with highest leverage while it is ranked 7946 by $\NM$. This pattern is extracted by SkOPUS for its leverage because this succession appears 174 times in the dataset, while its reversed pattern $\langle algorithm, paper\rangle$ appears only 80 times. It seems reasonable to be highly rank a pattern that occurs with more than twice the frequency of its reverse. $\NM$ ranks this pattern much lower because $paper$ and $algorithm$ are both probable individually (resp. support of 43\% and 58\%). A similar phenomenon is seen for $\langle base, result\rangle$ ---~which SkOPUS ranks  6 vs. 650 for $\NM$~--- and for $\langle learn, result\rangle$---~which SkOPUS ranks 8 vs. $25,810$ for $\NM$. %\fpcomment{@Nikolaj, is that a possible explanation?, sounds about right} 
\item Patterns $\NM$ ranks high but not SkOPUS: $\langle reproduc, hilbert\rangle$ is
extracted as the  7\textsuperscript{th} pattern by $\NM$ while it is ranked 696
by SkOPUS. This pattern is extracted by $\NM$ because it consists of two relatively rare items, \emph{reproduc} occurs 32 times and \emph{hilbert} occurs 36 times. 
On the other hand, SkOPUS ranks this pattern relatively low
because it appears in only 28 abstracts.  While under our measure of expected support this pattern has high lift, it has low leverage, the latter being a measure that favors patterns that appear the greatest number of times in excess of the expected. It is also interesting to note that SkOPUS ranks the pattern $\langle
reproduc, hilbert, spac\rangle$ much higher than $\langle reproduc,
hilbert\rangle$ (390 vs. 696). It seems natural that if all abstracts that
include $\langle reproduc, hilbert\rangle$ are followed by $space$, then the
pattern of length 3 should be more interesting. SkOPUS' rank respects this
ordering while $\NM$ does not in this case, having the pattern $\langle
reproduc, hilbert, spac\rangle$ ranked 42\textsuperscript{nd}. \rev{Note also that the pattern $\langle hilbert,spac\rangle$ occurs less frequently than $\langle reproduc, hilbert\rangle$, as some papers mention ``reproduc[ing] the Hilbert \textbf{norm}'', but not space (see e.g. the paper by F. Bach at \url{http://www.jmlr.org/papers/v9/bach08b.html}).}
\end{itemize}

\subsection{Execution time}
We finish the experiments by showing the running time of our approach; all experiments are performed using a standard desktop computer. 

Results are reported in Table~\ref{tab:runtime}. Not surprisingly, extracting the top-20 under support with SkOPUS is extremely fast, because the most frequent elements are encountered very early in the exploration of the search space;\footnote{Although $support$ is often not a good proxy for interestingness, it remains that SkOPUS can be very efficiently and exactly extract the most frequent sequential patterns. } no top-20 were actually presenting patterns of more than length 2 regardless of the experiments. 

For other methods, it can be observed than synthetic data are extremely
challenging, mostly because we included only a few patterns with reasonable
probability. Results are obtained in a few hours for SkOPUS under leverage and
less than a minute for $\NM$. The main reason why $\NM$ is faster is due to a
high mining threshold of 500 without which extraction cannot be performed, because if we lower this threshold then the mining step will take a considerate time due to a frequent pattern explosion. On the other hand, SkOPUS does not require any threshold. 

Finally, our approach exhibits extremely competitive running time compared to support on real-world datasets where interesting patterns are more present. It only takes SkOPUS 37s to extract the top-20 patterns in the JMLR dataset; only slightly more than twice the time taken for support-extraction, and significantly faster than $\NM$ with minimum support set to 10. On other literary works, SkOPUS finishes in less than a few minutes, while $\NM$ is generally finishes in seconds but it required to set a minimum support threshold of 5. 

It is finally interesting to note that the algorithmic complexity of SkOPUS and $\NM$ varies with different elements. For SkOPUS, it is a function of how interesting the patterns in the data are: data that holds patterns of high interest will prune significantly large parts of the search space (and the earlier they are found, the quicker the process). For $\NM$, the interests of the patterns that the data holds is almost neutral to the complexity, which will mostly vary as a function of the number of frequent closed patterns. 

\begin{table}
\centering
\caption{\label{tab:runtime}Running time on the different datasets for extracting the top-$20$.}
\begin{tabular}{lrrr}
\toprule
&  \multicolumn{3}{l}{Runtimes}\\[2pt]
\cmidrule{2-4}
Name &  Support & Leverage &$\NM$~\cite{Tatti2015-partition} (mining + ranking)\\[2pt]
\midrule

\emph{Synth-1} &$<1$s &
1h41m57s &
5s $+$ 4s
\\
\emph{Synth-2} &$<1$s & 
1h44m28s &
5s $+$ 5s
\\
\emph{Synth-3} &$<1$s &
53m19s &
9s $+$ 9s
\\
\emph{Synth-4} &$<1$s &
1h24m40s &
6s $+$ 6s
\\
\emph{Synth-5} &$<1$s &
1h34m41s &
7s $+$ 7s
\\
\emph{Synth-6} &$<1$s &
3h55m33s &
10s $+$ 11s
\\
\emph{Synth-7} &$<1$s &
6h5m31s &
18s $+$ 19s
\\
\emph{Synth-8} &$<1$s &
1h25m25s &
16s $+$ 14s
\\
\emph{Synth-9} &$<1$s &
2h28m30s &
21s $+$ 24s
\\
\emph{Synth-10} &$<1$s &
8h58m53s &
8m15s $+$ 6m16s
\\
\emph{JMLR} & 16s & 37s &
6m47s + 1m3s \\
\emph{Lawyers}
&26s&3m51s&
13s $+$ 2s
\\
\emph{Finn}
&6s&1m8s&
5s $+$ 2s
\\
\emph{Sawyer}
&7s&1m23s&
2s $+$ 1s
\\
\emph{NewPhysics}
&5s&48s&
3s $+$ 1s
\\
\emph{TwoCities}
&14s&2m16s&
6s $+$ 2s
\\
\emph{Animals}
&8s&1m46s&
1s $+$ 1s
\\
\bottomrule
\end{tabular}
\end{table}

\section{Conclusion}
This paper introduced a new general definition for the expected support of sequential patterns, which specifically focuses on the order within the pattern. 

We have described the intuition behind our definitions, as well as efficient algorithms for both the computation of the expected support and the exact exploration of the search space. 
Put together, these contributions allowed us to introduce SkOPUS which constitutes, to the best of our knowledge, the first framework for the exact mining of the $k$  sequential patterns with highest leverage from data. 

Experiments on controlled data have validated the consistency and relevance of our framework, relevance that we have then confirmed on literary work. 

This work naturally raised a number of questions and issues that, we anticipate, will be of great interest to the community in the future. These include: 
\begin{enumerate}
\item finding heuristics to quickly fill the temporary top-$k$ with patterns of high-interest, possibly by integrating some background knowledge;
\item refining upper bounds for the most common interestingness measures; 
\item filtering the results to remove trivial subpatterns from those that are discovered (patterns highlighted in boldface in Table~3); 
\item assessing the statistical significance of the extracted patterns, so that only patterns that have high-likelihood to be observed in future data would be extracted (as initiated by \cite{Castro2012significant,Achar2015}); 
\item using background knowledge to define a model of the joint distribution, and then extract the patterns that differ the most from it; such an approach has been investigated by S. Jaroszewicz on non-sequential patterns, and models the joint distribution with a Bayesian Network \cite{Szymon2004}; 
\item \rev{many domains hold patterns of the type ``admission, then biopsy or blood-test, and then surgery''. Our work focuses on the extraction of chains of events (serial episodes); we believe that extension to the more general classes of episodes will be of major interest. }
\end{enumerate}

\begin{acknowledgements}
We are grateful to the anonymous reviewers and to Matteo Riondato for valuable and constructive comments on this manuscript. 
\end{acknowledgements}

\section*{Supplementary material}
For completeness, we make available the top-100 patterns extracted from six literary works (see Table~\ref{tab:description-datasets}) and for all the methods discussed at \url{https://github.com/fpetitjean/Skopus/}.
\begin{table}
\centering
\caption{\label{tab:description-datasets}Characteristics of the literary datasets}
\begin{tabular}{lcccc}
\hline \\[-5pt]
Name &  $|\mathcal{I}|$ & $|\mathcal{D}|$ & Avg. length & Max. length \\[2pt]
\hline \\[-5pt]
\emph{The Industries of Animals} & 5547 & 4931 & 12.8 & 107\\[2pt]
\emph{A Book About Lawyers} & 8386 & 4787 & 21.9 & 211\\[2pt]
\emph{Adventures of Huckleberry Finn} & 4504 & 6402 & 13.7 & 213\\[2pt]
\emph{A Tale of Two Cities} & 6392 & 8584 & 12.4 &  150\\[2pt]
\emph{The Adventures of Tom Sawyers} & 5073 & 5259 & 10.7 &  119\\[2pt]
% \emph{File Transfer Protocol} &  1209 & 1337 & 10.2 &  125& 716&19.0 &  173\\[2pt]
% \emph{Simple Mail Transfer Protocol} &  1829 & 2226 & 11.4 &  61& 959&26.4 &  173\\[2pt]
% \emph{SSH Connection Protocol} &  893 &  533 &  9.9 &  63& 254& 20.8 &  236\\[2pt]
\emph{The New Physics and Its Evolution} &  3832 &  3081 &  19.4 &  81\\[1pt]
\hline
\end{tabular}
\end{table}

\section*{Compliance with Ethical Standards}
This material is based upon work supported by the Air Force Office of Scientific Research, Asian Office of Aerospace Research and
Development (AOARD) under award number FA2386-15-1-4007.
This research was also supported by China Special Fund for Meteorological Research in the Public Interest under grant GYHY201306070, Jiangsu Government Scholarship for Overseas Studies; and by the Australian Research Council under grant DP140100087.


\bibliographystyle{spmpsci}
\bibliography{sequencereferences}

\begin{thebibliography}{10}
\providecommand{\url}[1]{{#1}}
\providecommand{\urlprefix}{URL }
\expandafter\ifx\csname urlstyle\endcsname\relax
  \providecommand{\doi}[1]{DOI~\discretionary{}{}{}#1}\else
  \providecommand{\doi}{DOI~\discretionary{}{}{}\begingroup
  \urlstyle{rm}\Url}\fi

\bibitem{RefOECFactsEnglish}
{The Oxford English Corpus: Facts about Language}.
\newblock In: {Oxford Dictionaries}. {Oxford University Press} (2015).
\newblock
  \url{http://www.oxforddictionaries.com/words/the-oec-facts-about-the-language}

\bibitem{Achar2012}
Achar, A., Laxman, S., Viswanathan, R., Sastry, P.: Discovering injective
  episodes with general partial orders.
\newblock Data Mining and Knowledge Discovery \textbf{25}(1), 67--108 (2012)

\bibitem{Achar2015}
Achar, A., Sastry, P.: Statistical significance of episodes with general
  partial orders.
\newblock Information Sciences \textbf{296}(0), 175--200 (2015)

\bibitem{aggarwal:14:fpm}
Aggarwal, C.C., Han, J. (eds.): Frequent pattern mining.
\newblock Springer (2014)

\bibitem{AgrawalEtAl:1993}
Agrawal, R., Imielinski, T., Swami, A.: Mining association rules between sets
  of items in large databases.
\newblock In: Proceedings of the 1993 {ACM}-{SIGMOD} International Conference
  on Management of Data, pp. 207--216. Washington, DC (1993)

\bibitem{RefAgrawal1995}
Agrawal, R., Srikant, R.: Mining sequential patterns.
\newblock In: Proceedings of the Eleventh International Conference on Data
  Engineering, March 6-10, 1995, Taipei, Taiwan, pp. 3--14. {IEEE} Computer
  Society (1995)

\bibitem{bayardo1999mining}
Bayardo~Jr, R.J., Agrawal, R.: Mining the most interesting rules.
\newblock In: Proceedings of the fifth ACM SIGKDD international conference on
  Knowledge discovery and data mining, pp. 145--154. ACM (1999)

\bibitem{RefBoley2009}
Boley, M., Horv{\'{a}}th, T., Wrobel, S.: Eficient discovery of interesting
  patterns based on strong closedness.
\newblock Statistical Analysis and Data Mining \textbf{2}(5-6), 346--360 (2009)

\bibitem{Castro2012significant}
Castro, N.C., Azevedo, P.J.: Significant motifs in time series.
\newblock Statistical Analysis and Data Mining: The ASA Data Science Journal
  \textbf{5}(1), 35--53 (2012)

\bibitem{RefFournier2013}
Fournier{-}Viger, P., Gomariz, A., Gueniche, T., Mwamikazi, E., Thomas, R.:
  {TKS:} efficient mining of top-k sequential patterns.
\newblock In: Advanced Data Mining and Applications, 9th International
  Conference, {ADMA} 2013, Hangzhou, China, December 14-16, 2013, Proceedings,
  Part {I}, \emph{Lecture Notes in Computer Science}, vol. 8346, pp. 109--120.
  Springer (2013)

\bibitem{geng2007choosing}
Geng, L., Hamilton, H.J.: Choosing the right lens: Finding what is interesting
  in data mining.
\newblock In: Quality Measures in Data Mining, pp. 3--24. Springer (2007)

\bibitem{Gwadera2005-markov}
Gwadera, R., Atallah, M.J., Szpankowski, W.: Markov models for identification
  of significant episodes.
\newblock In: SIAM International Conference on Data Mining, pp. 404--414 (2005)

\bibitem{Gwadera2005-independence}
Gwadera, R., Atallah, M.J., Szpankowski, W.: Reliable detection of episodes in
  event sequences.
\newblock Knowledge and Information Systems \textbf{7}(4), 415--437 (2005)

\bibitem{RefGwadera2010}
Gwadera, R., Crestani, F.: Ranking sequential patterns with respect to
  significance.
\newblock In: Advances in Knowledge Discovery and Data Mining, 14th
  Pacific-Asia Conference, {PAKDD} 2010, Hyderabad, India, June 21-24, 2010.
  Proceedings. Part {I}, \emph{Lecture Notes in Computer Science}, vol. 6118,
  pp. 286--299. Springer (2010)

\bibitem{RefHam2010}
H{\"{a}}m{\"{a}}l{\"{a}}inen, W.: Efficient discovery of the top-k optimal
  dependency rules with {Fisher}'s exact test of significance.
\newblock In: {ICDM} 2010, The 10th {IEEE} International Conference on Data
  Mining, Sydney, Australia, 14-17 December 2010, pp. 196--205. {IEEE} Computer
  Society (2010)

\bibitem{RefHan2007}
Han, J., Cheng, H., Xin, D., Yan, X.: Frequent pattern mining: current status
  and future directions.
\newblock Data Mining and Knowledge Discovery \textbf{15}(1), 55--86 (2007)

\bibitem{RefHan2000}
Han, J., Pei, J., Mortazavi{-}Asl, B., Chen, Q., Dayal, U., Hsu, M.: Freespan:
  frequent pattern-projected sequential pattern mining.
\newblock In: Proceedings of the sixth {ACM} {SIGKDD} international conference
  on Knowledge discovery and data mining, Boston, MA, USA, August 20-23, 2000,
  pp. 355--359. {ACM} (2000)

\bibitem{Szymon2004}
Jaroszewicz, S., Simovici, D.A.: Interestingness of frequent itemsets using
  bayesian networks as background knowledge.
\newblock In: Proceedings of the Tenth ACM SIGKDD International Conference on
  Knowledge Discovery and Data Mining, pp. 178--196. ACM (2004)

\bibitem{Lam2014mining-compressing}
Lam, H.T., Moerchen, F., Fradkin, D., Calders, T.: Mining compressing
  sequential patterns.
\newblock Statistical Analysis and Data Mining: The ASA Data Science Journal
  \textbf{7}(1), 34--52 (2014)

\bibitem{RefLowKam2013}
Low{-}Kam, C., Ra{\"{\i}}ssi, C., Kaytoue, M., Pei, J.: Mining statistically
  significant sequential patterns.
\newblock In: 2013 {IEEE} 13th International Conference on Data Mining, Dallas,
  TX, USA, December 7-10, 2013, pp. 488--497. {IEEE} Computer Society (2013)

\bibitem{mabroukeh2010taxonomy}
Mabroukeh, N.R., Ezeife, C.I.: A taxonomy of sequential pattern mining
  algorithms.
\newblock ACM Computing Surveys (CSUR) \textbf{43}(1), 3 (2010)

\bibitem{RefMampaey2012}
Mampaey, M., Vreeken, J., Tatti, N.: Summarizing data succinctly with the most
  informative itemsets.
\newblock ACM Transactions on Knowledge Discovery from Data \textbf{6}(4), 16
  (2012)

\bibitem{RefMannila1995}
Mannila, H., Toivonen, H., Verkamo, A.I.: Discovering frequent episodes in
  sequences.
\newblock In: Proceedings of the First International Conference on Knowledge
  Discovery and Data Mining (KDD-95), Montreal, Canada, August 20-21, 1995, pp.
  210--215. {AAAI} Press (1995)

\bibitem{RefMannila1997}
Mannila, H., Toivonen, H., Verkamo, A.I.: Discovery of frequent episodes in
  event sequences.
\newblock Data Mining and Knowledge Discovery \textbf{1}(3), 259--289 (1997)

\bibitem{RefMasseglia1998}
Masseglia, F., Cathala, F., Poncelet, P.: The {PSP} approach for mining
  sequential patterns.
\newblock In: Principles of Data Mining and Knowledge Discovery, Second
  European Symposium, {PKDD} '98, Nantes, France, September 23-26, 1998,
  Proceedings, \emph{Lecture Notes in Computer Science}, vol. 1510, pp.
  176--184. Springer (1998)

\bibitem{mooney2013sequential}
Mooney, C.H., Roddick, J.F.: Sequential pattern mining--approaches and
  algorithms.
\newblock ACM Computing Surveys (CSUR) \textbf{45}(2), 19 (2013)

\bibitem{RefPei2001}
Pei, J., Han, J., Mortazavi{-}Asl, B., Pinto, H., Chen, Q., Dayal, U., Hsu, M.:
  Prefixspan: Mining sequential patterns by prefix-projected growth.
\newblock In: Proceedings of the 17th International Conference on Data
  Engineering, April 2-6, 2001, Heidelberg, Germany, pp. 215--224. {IEEE}
  Computer Society (2001)

\bibitem{piatetsky:daapos}
Piatetsky-Shapiro, G.: Discovery, analysis, and presentation of strong rules.
\newblock In: G.~Piatetsky-Shapiro, J.~Frawley (eds.) Knowledge Discovery in
  Databases, pp. 229--248. AAAI/MIT Press, Menlo Park, CA. (1991)

\bibitem{RefRaissi2008}
Ra{\"{\i}}ssi, C., Calders, T., Poncelet, P.: Mining conjunctive sequential
  patterns.
\newblock Data Mining and Knowledge Discovery \textbf{17}(1), 77--93 (2008)

\bibitem{RefSiebes2006}
Siebes, A., Vreeken, J., van Leeuwen, M.: Item sets that compress.
\newblock In: Proceedings of the Sixth {SIAM} International Conference on Data
  Mining, April 20-22, 2006, Bethesda, MD, {USA}, pp. 395--406. {SIAM} (2006)

\bibitem{Tatti2009}
Tatti, N.: Significance of episodes based on minimal windows.
\newblock In: IEEE International Conference on Data Mining, pp. 513--522 (2009)

\bibitem{Tatti2014-compact-windows}
Tatti, N.: Discovering episodes with compact minimal windows.
\newblock Data Mining and Knowledge Discovery \textbf{28}(4), 1046--1077 (2014)

\bibitem{Tatti2015-partition}
Tatti, N.: Ranking episodes using a partition model.
\newblock Data Mining and Knowledge Discovery pp. 1--31 (2015)

\bibitem{RefTatti2010}
Tatti, N., Mampaey, M.: Using background knowledge to rank itemsets.
\newblock Data Mining and Knowledge Discovery \textbf{21}(2), 293--309 (2010)

\bibitem{Tatti2012}
Tatti, N., Vreeken, J.: The long and the short of it: Summarising event
  sequences with serial episodes.
\newblock In: Proceedings of the Tenth ACM SIGKDD International Conference on
  Knowledge Discovery and Data Mining, pp. 462--470 (2012)

\bibitem{RefTew2014}
Tew, C.V., Giraud{-}Carrier, C.G., Tanner, K.W., Burton, S.H.: Behavior-based
  clustering and analysis of interestingness measures for association rule
  mining.
\newblock Data Mining and Knowledge Discovery \textbf{28}(4), 1004--1045 (2014)

\bibitem{Tucker2006}
Tucker, A.: Applied Combinatorics.
\newblock John Wiley \& Sons, Inc., New York, NY, USA (2006)

\bibitem{RefTzvetkov2003}
Tzvetkov, P., Yan, X., Han, J.: {TSP:} mining top-k closed sequential patterns.
\newblock In: Proceedings of the 3rd {IEEE} International Conference on Data
  Mining {(ICDM} 2003), 19-22 December 2003, Melbourne, Florida, {USA}, pp.
  347--354. {IEEE} Computer Society (2003)

\bibitem{RefVreeken2011}
Vreeken, J., van Leeuwen, M., Siebes, A.: Krimp: mining itemsets that compress.
\newblock Data Mining and Knowledge Discovery \textbf{23}(1), 169--214 (2011)

\bibitem{RefWebb2010}
Webb, G.: Self-sufficient itemsets: An approach to screening potentially
  interesting associations between items.
\newblock Transactions on Knowledge Discovery from Data \textbf{4}, 3:1--3:20
  (2010)

\bibitem{RefWebb1995}
Webb, G.I.: {OPUS:} an efficient admissible algorithm for unordered search.
\newblock Journal of Artificial Intelligence Research {(JAIR)} \textbf{3},
  431--465 (1995)

\bibitem{webb2000efficient}
Webb, G.I.: Efficient search for association rules.
\newblock In: Proceedings of the Sixth ACM SIGKDD International Conference on
  Knowledge Discovery and Data Mining, pp. 99--107. ACM (2000)

\bibitem{RefWebb2007}
Webb, G.I.: Discovering significant patterns.
\newblock Machine Learning \textbf{68}(1), 1--33 (2007)

\bibitem{RefWebb2008}
Webb, G.I.: Layered critical values: {A} powerful direct-adjustment approach to
  discovering significant patterns.
\newblock Machine Learning \textbf{71}(2-3), 307--323 (2008)

\bibitem{Webb2010}
Webb, G.I.: Self-sufficient itemsets: An approach to screening potentially
  interesting associations between items.
\newblock ACM Transactions on Knowledge Discovery from Data (TKDD)
  \textbf{4}(1), 3 (2010)

\bibitem{RefWebb2011}
Webb, G.I.: Filtered-top-\emph{k} association discovery.
\newblock Wiley Interdisc. Rev.: Data Mining and Knowledge Discovery
  \textbf{1}(3), 183--192 (2011)

\bibitem{RefWebb2014}
Webb, G.I., Vreeken, J.: Efficient discovery of the most interesting
  associations.
\newblock ACM Transactions on Knowledge Discovery from Data \textbf{8}(3),
  1--31 (2014)

\bibitem{RefYan2006}
Yan, X., Han, J., Afshar, R.: Clospan: Mining closed sequential patterns in
  large databases.
\newblock In: Proceedings of the Third {SIAM} International Conference on Data
  Mining, San Francisco, CA, USA, May 1-3, 2003, pp. 166--177. {SIAM} (2003)

\bibitem{RefZimmermann2013}
Zimmermann, A.: Objectively evaluating interestingness measures for frequent
  itemset mining.
\newblock In: J.~Li, L.~Cao, C.~Wang, K.~Tan, B.~Liu, J.~Pei, V.~Tseng (eds.)
  Trends and Applications in Knowledge Discovery and Data Mining, \emph{Lecture
  Notes in Computer Science}, vol. 7867, pp. 354--366. Springer Berlin
  Heidelberg (2013)

\end{thebibliography}

\end{document}